\documentclass[reqno]{amsart}
\usepackage{mathtools}

\usepackage[dvipsnames]{xcolor}
\usepackage{hyperref}
\definecolor{burntorange}{HTML}{BF5700}
\definecolor{UTblue}{HTML}{00A9B7}
\definecolor{bluebonnet}{HTML}{005F86}
\hypersetup{
    pdftitle={Geometric structure of shallow neural networks},
    linktoc=all,     %set to all if you want both sections and subsections linked
    colorlinks=true, %set true if you want colored links
    linkcolor=bluebonnet,
    citecolor=burntorange,
    urlcolor=bluebonnet,
}

\def\tr{{\rm Tr}}

\def\N{{\mathbb N}}

\def\R{{\mathbb R}}

\def\cC{{\mathcal C}}

\def\cF{{\mathcal F}}

\def\cL{{\mathcal L}}

\def\cN{{\mathcal N}}
\def\cP{{\mathcal P}}
\def\cR{{\mathcal R}}

\def\diag{{\rm diag}}

\def\rank{{\rm rank}}
\def\Pen{{\rm Pen}}

\def\CostN{\cC_{\cN}}
\newcommand{\norm}[1]{\ensuremath{\left\lVert{#1}\right\rVert}}
\newcommand{\abs}[1]{\ensuremath{\left\lvert{#1}\right\rvert}}

\def\1{{\bf 1}}

\def\argmin{{\rm argmin}}
\def\ran{{\rm range}}
\def\X0{X_0}

\def\eqnn{\begin{eqnarray*}}
\def\eeqnn{\end{eqnarray*}}
\def\eqn{\begin{eqnarray}}
\def\eeqn{\end{eqnarray}}

\def\prf{\begin{proof}}
\def\endprf{\end{proof}}

\theoremstyle{plain}
\newtheorem{theorem}{Theorem}[section]
\newtheorem{definition}[theorem]{Definition}

\newtheorem{remark}[theorem]{Remark}

\numberwithin{equation}{section}

\begin{document} 

\title[Geometric Structure of Shallow Neural Networks]
{Geometric structure of shallow neural networks and constructive $\cL^2$ cost minimization}

\author{Thomas Chen}
\address[T. Chen]{Department of Mathematics, University of Texas at Austin, Austin TX 78712, USA}
\email{tc@math.utexas.edu} 
\author{Patr\'{i}cia Mu\~{n}oz Ewald}
\address[P. M. Ewald]{Department of Mathematics, University of Texas at Austin, Austin TX 78712, USA}
\email{ewald@utexas.edu} 

\begin{abstract}
    In this paper, we approach the problem of cost (loss) minimization in
    underparametrized shallow ReLU networks through the explicit construction of upper
    bounds which appeal to the structure of classification data, 
    without use of gradient descent. A key focus is on elucidating the
    geometric structure of approximate and precise minimizers. 
    We consider 
    %shallow neural networks with ReLU activation function, 
    an ${\mathcal L}^2$
    cost function, 
    input space ${\mathbb R}^M$, output space ${\mathbb R}^Q$ with $Q\leq M$, 
    and training input sample size 
    that can be arbitrarily large. 
    We prove an upper bound on the minimum of the cost
    function of order $O(\delta_P)$ where $\delta_P$ measures the signal-to-noise ratio of
    training data. In the special case $M=Q$, we explicitly determine an exact
    degenerate local minimum of the cost function, and show that the sharp value differs
    from the upper bound obtained for $Q\leq M$ by a relative error $O(\delta_P^2)$. The
    proof of the upper bound yields a constructively trained network; we show that it
    metrizes a particular $Q$-dimensional subspace in the input space ${\mathbb R}^M$.  
    We comment on the characterization of the global minimum of the cost function in the
    given context.  
\end{abstract}

\maketitle

\section{Introduction}

Applications of neural networks have rapidly become indispensable in a
vast range of research disciplines, and their technological impact in recent years has
been enormous. However, despite those successes, the fundamental conceptual reasons
underlying their functioning are currently insufficiently understood, and remain the
subject of intense investigation.  
The common approach to the
minimization of the cost function is based on the  gradient flow generated
by it, and a variety of powerful algorithms have been developed for this purpose that can
be highly successful in suitable application scenarios, see e.g. \cite{lcbh, arora-ICM-1,
grokut-1}; nevertheless, even if a global or sufficiently good local minimum is obtained,
the properties of the minimizing
parameters (weights, biases) in general remain quite mysterious, see e.g., 
\cite{hanrol, nonreeste}.

% The current paper is the first in a series of works in which we investigate the geometric
% structure of cost minimizers in neural networks (that is, weights and biases that minimize
% the cost). 
% Here, we analyze shallow neural networks in the context of supervised learning
% for multiclass classification, with one hidden layer, ReLU activation function, an $\cL^2$
% Schatten class (or Hilbert-Schmidt) cost function, input space ${\mathbb R}^M$,   output
% space ${\mathbb R}^Q$ with $Q\leq M$, and training input sample size $N>QM$ that can be
% arbitrarily large---therefore, we are considering the underparametrized regime. 

For linear neural networks and the $\mathcal{L}^{2}$ cost, critical points are well
understood. Consider a matrix $X \in \mathbb{R}^{M \times N}$, whose columns are made up
from the vectors in a training data set, and a matrix $Y \in \mathbb{R}^{Q \times N}$
whose columns are the corresponding target vectors. It is well known 
that a solution to
\begin{align}
    \min_{W} \|WX - Y\|_{\mathcal{L}^{2}}
\end{align}
is given by $W = Y \,\Pen[X]$, where $\Pen[X]$ is the Penrose inverse or pseudoinverse of
$X$.

It is helpful to think in terms of projections. Note that 
$\Pen[X]X$ is a projector onto
the orthogonal complement of the kernel of $X$, $\Pen[X]X = \mathcal{P}_{(\ker
X)^{\perp}}$. Therefore, another way of stating the solution to the linear regression
problem is
\begin{align}
    \label{linear-bound1}
    \min_{W} \|WX - Y\|_{\mathcal{L}^{2}} = 
    \norm{Y \Pen[X]X - Y}_{\mathcal{L}^{2}} =
    \|Y \mathcal{P}_{(\ker
    X)^{\perp}}^{\perp}\|_{\mathcal{L}^{2}}.
\end{align}
One can bound this as follows
\begin{align}
    \label{linear-bound2}
    \norm{Y \mathcal{P}_{\ker X}}_{\mathcal{L}^{2}} \leq \norm{Y}_{op} \sqrt{\dim \ker X}
    = \norm{Y}_{op} \sqrt{N - \rank X}.
\end{align}
In the overdetermined case $M \geq N$, when $X$ is full-rank the cost can be taken
to zero even with a linear network. 
We assume here that $N$ can be arbitrarily large, so
that a zero-loss global minimum is not guaranteed by \eqref{linear-bound2}.

One of the goals of introducing non-linearities is to increase the complexity of the
model. In this work, we study shallow neural networks
\begin{align}
    X^{(2)} = W_2 X^{(1)} + B_2 \in \mathbb{R}^{Q\times N}, \quad X^{(1)} = \sigma(W_1 X +
    B_1) \in \mathbb{R}^{M\times N},
\end{align}
where $\sigma$ is the source of non-linearity, called an activation function.
The ReLU activation function is a popular choice in applications; it is defined on vectors
component-wise as the ramp function $(a)_{+} = \max\{0,a\}$.
Because it acts as the identity function on vectors with non-negative components, it is
convenient to study subcases where the network becomes linear so that results for linear
networks apply, and this has been the subject of a few works \cite{kawaguchi16,
zhouliang18}.

Minimizing the $\mathcal{L}^{2}$ cost with respect to $W_2$ yields
\begin{align}
    \min_{W_2} \norm{W_2 X^{(1)} + B_2 - Y}_{\mathcal{L}^{2}} &= \norm{(Y-B_2)
    \Pen[X^{(1)}]X^{(1)} - (Y-B_2)}_{\mathcal{L}^{2}} \nonumber \\
                                            &= \norm{(Y-B_2) \mathcal{P}_{\ker
                                            X^{(1)}}}_{\mathcal{L}^{2}}.
\end{align}
The rank of $\mathcal{P}_{\ker X^{(1)}}$ is influenced by the coordinates
picked out by $W_1$ and the ReLU activation function. 
When working with classification, there is a natural choice of coordinate frame: Suppose
the set of training data is split into $Q$ classes. Then we may consider the mean of each
class, and these vectors form the columns of a matrix $\overline{X_{0}^{red}} \in
\mathbb{R}^{M\times Q}$. The deviation of each vector in the training data set from the
corresponding target class mean is stored in the matrix $\Delta X_{0}$. In Theorem
\ref{thm-cC-uppbd-1}, we derive an upper bound on the minimum of the cost function with an
explicit construction of suitable weights and biases:
\begin{align}
    \label{bound}
    \min_{W_1, b_1, W_2, b_2} \norm{W_2 X^{(1)} + B_2 - Y}_{\mathcal{L}^{2}} 
    &\leq C \, \|Y \Pen[ \overline{X_{0}^{red}}] \Delta X_{0} \|_{\mathcal{L}^{2}}.
   %  \nonumber \\
   %  &\leq C \, \|Y\|_{op} \, \delta_{P}.
\end{align}
Define the parameter
\begin{align}
    \delta_{P} := \sup_{\Delta x_0} \abs{\Pen[\overline{X_{0}^{red}}] \Delta x_{0}},
\end{align}
which measures the relative size between the averages in $\overline{X_{0}^{red}}$ and the
deviations in $\Delta X_{0}$. Then \eqref{bound} can be bounded in terms of this
signal-to-noise ratio,
\begin{align}
    \min_{W_1, b_1, W_2, b_2} \norm{W_2 X^{(1)} + B_2 - Y}_{\mathcal{L}^{2}} 
     &\leq C \, \|Y\|_{op} \, \delta_{P}.
\end{align}

The proof makes crucial use of the bias terms, where $b_1$ is constructed so that the
terms in the span of the averages survive $\sigma$, and the terms in the complement
disappear.  
When analyzing the particular case $M=Q$, the same strategy creates parameters which
result in 
a linear neural network.  Nevertheless, it is still possible to extract a meaningful
bound:  in Theorem \ref{thm-cC-uppbd-2}, we explicitly
determine an exact degenerate local minimum of the cost function, and show that the sharp
value differs from the upper bound \eqref{bound} obtained for $Q\leq M$ by a relative error
$O(\delta_P^2)$.

In Theorem \ref{thm-DL-geometry-1}, we use the parameters given in Theorem
\ref{thm-cC-uppbd-1} to define a metric in (a subspace of) input space, and show that the
problem of classifying multiclass data is equivalent to a metric minimization problem.

In Theorem \ref{thm-CostN-truc-1-0}, we prove a result for $Q=M$ closely related to
Theorem \ref{thm-cC-uppbd-2}, but circumventing the need for the network to be linear, via
the introduction of a truncation map. 

In Section \ref{experiments}, we compare our theoretical bounds with numerical results
based on randomly initialized neural networks trained with gradient descent on synthetic
data.  We show that the
bound improves with the increase of clustering of the data, whereby $\delta_{P}$
decreases; this matches our prediction.

Finally, we comment on the assumptions: We study shallow neural networks with 
dimensions of input, hidden and output layer $(M, M, Q)$, respectively, where $M\geq Q$
and at one point specialize to $M=Q$. 
While this is restrictive in the context of general
neural networks, this can be useful as the study of the last layers of a deep network.
% Moreover, the current paper is the first in a series of works in which we investigate the
% geometric structure of cost minimizers in neural networks, including deep networks with
% more general architectures. 

Moreover, we place no restriction on the number of training data
points $N$ compared with the number of parameters, and so the results here hold for both
underparametrized and overparametrized networks.
In the strongly overparametrized regime $M>N$, every
data point is its own cluster, so that the cluster variances are identically zero.
Therefore, our construction yields zero training loss. This situation and its extension to
deep networks is studied in detail in \cite{bccem25}.

\subsection{Related work}
Early work on the loss landscape of neural networks already focused on shallow
architectures. For linear networks, \cite{baldihornik89} analyzed the square loss for
one-hidden-layer linear networks and showed that, under mild assumptions, every local
minimum is global.
\cite{kawaguchi16} extended this to deep linear networks, proving the absence of bad local
minima and
characterizing saddle points, and subsequent works refined the
description of critical points and their geometry, including explicit parametrizations and
second-order analyses \cite{zhouliang18,trageretal20,achouretal24}. For shallow ReLU
networks, the loss landscape can be partly reduced to a linear problem when biases are
ignored, and this reduction was exploited in \cite{kawaguchi16,zhouliang18}. \cite{cooper21}
instead made crucial use of bias vectors to construct global minima that lie outside a
fixed linear region for overparameterized networks, thereby showing that
genuinely nonlinear structure in the loss landscape is already present in the shallow case.

Despite the dominance of very deep architectures in applications, shallow
networks remain a central object in contemporary theory. From the optimization and
generalization side, mean-field analyses for two-layer networks describe gradient-based
training via limiting partial differential equations and yield convergence and generalization
guarantees, e.g. \cite{meietal18,meietal19}. In parallel, a
substantial literature studies the implicit bias of gradient descent and gradient flow,
both in regression and classification, e.g. \cite{soudryetal18,
gunasekaretal18, nitandaetal19, lyuli20,jinmontufar23}.

Shallow models also appear implicitly when one “freezes” all but the last few layers of a deep
network. 
% In this setting, the last linear layer on top of learned features is a
% shallow network whose optimization can often be analyzed more cleanly.
This perspective underlies work on linear classifier probes, where linear predictors are
trained on top of intermediate representations to study how separability emerges across
layers \cite{alainbengio16}, and it is also closely connected to the unconstrained
features model (UFM) and the phenomenon of neural collapse. 
In the UFM, one replaces the
learned features by free variables and optimizes the last-layer problem jointly over
logits and features, reducing a deep model to a structured shallow one
\cite{zhuetal21, ewojt22, mixonetal24,tirerbruna22,hongling24}. 
In all these settings, shallow architectures serve both
as analytically tractable proxies and as faithful models of the final stages of deep
networks.

Finally, shallow networks are also a minimal setting in which one can contrast
lazy-training (kernel-like) and feature-learning regimes. In the infinite-width limit with
appropriate scaling, gradient descent on two-layer ReLU networks evolves essentially as a
linear model around initialization and leaves the representation almost fixed, the
so-called lazy-training regime \cite{jacotetal18,leeetal19,chizatbach19}. By changing
width, initialization scale, or training time one exits this regime and enters a genuinely
nonlinear feature-learning regime where the representation itself evolves during training
\cite{luoetal21}. Together with
mean-field analyses that capture representation dynamics,
this shows that shallow ReLU networks alone realize both extremes and the interpolation
between them \cite{montanariurbani25}, while remaining simple enough that their loss
landscape and implicit bias can be analyzed in detail.

\subsection{Related work by the authors}
We present the explicit construction of local and global $\cL^2$ cost minimizers in
deep neural networks 
for  clustered data in a separate work
\cite{chenewald23deep}, which invokes results developed in the paper at hand, combined
with inspiration drawn from rigorous renormalization group analysis in quantum field
theory, as for instance in \cite{bcfs-1}. 
In \cite{chenewald25JMLR}, we extend this construction to sequentially linearly separable
data sets which are significantly less structured than in \cite{chenewald23deep}.

In Section \ref{ssec-truncation}, we address the structure of
critical sets for shallow networks.  More generally, the comparison of constructive
training error bounds with those obtained from gradient descent is an interesting
theoretical problem.  A detailed analysis of the dynamical systems aspects of gradient
descent and its geometric interpretation for both shallow and deep networks is given in
\cite{chen24gradflow,chenewald24homotopy}.

\begin{remark}
    The inspiration for our approach is drawn from methods in mathematical physics
    aimed at determining the ground state energy for complex manybody quantum systems
    (such as large molecules), see for instance \cite{liesei}.
\end{remark}

\subsection{Outline of paper}

This paper is organized as follows. In Section \ref{sec-DL-def-1}, we give a detailed
description of the mathematical model and introduce notation.
% describing the shallow network, the cost function
% used and other notation. 
In Section \ref{sec-MainRes-1}, we present the main results, namely Theorem
\ref{thm-cC-uppbd-1}, which we prove in Section \ref{sec-prf-thm-cC-uppbd-1}, Theorem
\ref{thm-cC-uppbd-2}, which we prove in Section \ref{sec-prf-thm-cC-uppbd-2}, Theorem
\ref{thm-DL-geometry-1}, which we prove in Section \ref{sec-prf-thm-DL-geometry-1}, and
Theorem \ref{thm-CostN-truc-1-0}. 
In Section \ref{experiments}, we present some numerical experiments.
%In Section \ref{sec-DL-gradient-1}, we discuss a numerical
%example based on a gradient descent method.  
% In Section \ref{sec-theor-DL-1}, we provide a
% detailed description of the constructively trained network.

%%
\section{Definition of the mathematical model}
\label{sec-DL-def-1}

In this section, we give a detailed introduction of the mathematical model in discussion,
consisting of a shallow neural network for a classification task with $Q$ classes. We also
define relevant notation.

\vspace{1ex}
We introduce weight matrices 
\eqn 
	W_1 \in \R^{M\times M} 
	\;\;,\;\;
	W_2\in \R^{Q\times M}
\eeqn
and bias matrices
\eqn
	B_1 &:=& b_1u_N^T= [b_1\cdots b_1] \in \R^{M\times N}
	\nonumber\\
	B_2 &:=& b_2u_N^T=[b_2\cdots b_2] \in \R^{Q\times N}
\eeqn
where $b_1\in\R^M$ and $b_2\in\R^{Q}$ are column vectors, and 
\eqn 
	u_N = (1,1,\dots,1)^T \;\;\in\R^N \,.
\eeqn
Moreover, we choose to consider the ReLU activation function  
\eqn
	\sigma:\R^{M\times N}&\rightarrow&\R_+^{M\times N}
	\nonumber\\
	A=[a_{ij}] &\mapsto& [(a_{ij})_+]
\eeqn 
where 
\eqn\label{eq-rampfct=def-1}
	(a)_+:=\max\{0,a\}
\eeqn 
is the standard ramp function.
Then we define the neural network as follows: for a matrix of inputs $X_0 \in
\mathbb{R}^{M \times N}$,
\eqn \label{neuralnet1}
	X^{(1)} := \sigma(W_1 X_0 + B_1) \in \R_+^{M\times N}
\eeqn 
is the hidden layer, and 
\eqn \label{neuralnet2}
	X^{(2)} := W_2 X^{(1)} + B_2 \in \R^{Q\times N} \,
\eeqn 
is the terminal layer.

\vspace{1ex}
As the labels for the $Q$ classes, we consider a family $\{y_{j}\}_{j=1}^{Q}$ of
linearly independent target vectors, and define a target matrix
\eqn
	Y:=[y_1,\dots,y_Q] \in \R^{Q\times Q},
\eeqn 
which is invertible.
We also define the extension of the matrix of labels, $Y^{ext}$, by
\eqn 
	Y^{ext}:= [Y_1 \cdots Y_Q] \in \R^{Q\times N}
\eeqn 
where 
\eqn
	Y_j := [y_j \cdots y_j] \in \R^{Q\times N_j}
\eeqn
with $N_j$ copies of the same target column vector $y_j$.
Clearly, $Y^{ext}$ has full rank $Q$.

We let $\R^M$ denote the input space with coordinate system defined by  the orthonormal
basis vectors $\{e_\ell=(0,\dots,0,1,0,\dots,0)^T\in\R^M\,|\,\ell=1,\dots,M\}$.  For
$j\in\{1,\dots,Q\}$, and $N_j\in\N$, 
\eqn\label{eq-x0ji-def-1}
	x_{0,j,i}\in\R^M
	\;\;,\;\;
	i\in\{1,\dots,N_j\}
\eeqn 
denotes the $i$-th training input vector corresponding to the $j$-th target vector $y_j$. 
%We assume that all input vectors are contained in the positive quadrant, $\R_+^M$. 
We define the matrix of all training inputs belonging to $y_j$ by
\eqn 
	X_{0,j} := [x_{0,j,1} \cdots x_{0,j,i} \cdots x_{0,j,N_j} ] \,.
\eeqn  
% Correspondence of a training input vector to the same output $y_j$ defines an equivalence
% relation, that is, for each $j\in\{1,\dots,Q\}$ we have $x_{0,j,i}\sim x_{0,j,i'}$ for any
% $i,i'\in\{1,\dots,N_j\}$. Accordingly, $X_{0,j}$ labels the equivalence class of all
% inputs belonging to the $j$-th output $y_j$.
Then, we define the  matrix of training inputs
\eqn 
	\X0 := [X_{0,1}\cdots X_{0,j}\cdots X_{0,Q}] \in \R^{M\times N}
\eeqn 
where $N:=\sum_{j=1}^Q N_j$.

Let 
\eqn 
	\overline{x_{0,j}}:= \frac{1}{N_j}\sum_{i=1}^{N_j} x_{0,j,i} \in \R^M
\eeqn
denote the average over the equivalence class of all training input vectors corresponding
to the $j$-th class, and
\eqn
	\Delta x_{0,j,i}:= x_{0,j,i}-\overline{x_{0,j}}\,.
\eeqn 
Moreover, let
\eqn 
	\overline{X_{0,j}} := [\overline{x_{0,j}}\cdots \overline{x_{0,j}}] \in \R^{M\times N_j}
\eeqn 
and 
\eqn\label{eq-ovlnX0-def-1}
	\overline{X_0} := [\overline{X_{0,1}}\cdots \overline{X_{0,Q}}] \in \R^{M\times N} \,.
\eeqn 
Similarly, let
\eqn 
	\Delta X_{0,j} := [\Delta x_{0,j,1}\cdots \Delta x_{0,j,N_j}] \in \R^{M\times N_j} \,.
\eeqn
and
\eqn 
	\Delta X_{0} := [\Delta X_{0,1}\cdots \Delta X_{0,Q}] \in \R^{M\times N} \,.
\eeqn
Then, 
\eqn 
	X_{0} = \overline{X_0} + \Delta X_{0} \,,
\eeqn 
and we let
\eqn\label{eq-delt-1}
	\delta := \sup_{j,i}|\Delta x_{0,j,i}| \ \,.
\eeqn  

Next, we define the reduction of $\overline{X_0}$ as
\eqn
	\overline{X_0^{red}} := [\overline{x_{0,1}}\cdots \overline{x_{0,Q}}] \in \R^{M\times Q}
\eeqn 
and assume that its rank is maximal, $\rank(\overline{X_0^{red}})=Q$. The latter condition
is necessary for the averages of training inputs to be able to distinguish between the
different target values $y_j$, $j=1,\dots,Q$. 
Moreover, in this case the matrix
\eqn
	(\overline{X_0^{red}})^T\overline{X_0^{red}} \in \R^{Q\times Q}
\eeqn
is invertible, and we can define the Penrose inverse\footnote{Also called Moore--Penrose
inverse or pseudoinverse and denoted $A^{+}$ instead of $\Pen[A]$.} of
$\overline{X_0^{red}}$,
\begin{align}
    \label{penrose-inverse}
	\Pen[\overline{X_0^{red}}]
	:= \big((\overline{X_0^{red}})^T\overline{X_0^{red}}\big)^{-1}
	(\overline{X_0^{red}})^T \in \R^{Q\times M}.
\end{align}
The orthoprojector onto the range of $\overline{X_0^{red}}$ is then given by
\begin{align}
    \label{orthoprojector}
	P:= \overline{X_0^{red}} 
	\Pen[\overline{X_0^{red}}]
	\in \R^{M\times M}
\end{align}
and we note that $\Pen[\overline{X_0^{red}}]\overline{X_0^{red}}=\1_{Q\times Q}$. The
projector property $P^2=P$ is thus easily verified, and orthogonality with respect to the
Euclidean inner product $(\cdot,\cdot )$ on $\R^M$ holds due to $P^T=P$, whereby
$(v,Pw)=(Pv,w)$ for any $v,w\in\R^M$.  In particular, we have
\eqn 
	P\overline{X_0^{red}} = \overline{X_0^{red}}
	\;\;\;{\rm and}\;\;\;
	P\overline{X_0} = \overline{X_0} \,,
\eeqn 
%and
%\eqn 
%	\overline{x} = P\overline{x} \,,
%\eeqn
by construction.

\vspace{1ex} 
As the objective function, we consider the $\cL^2$ Schatten class (or
Hilbert-Schmidt) cost function  
\begin{align}
\label{eq-cC-def-1}
    \cC[W_j,b_j] &:= \frac1{\sqrt N}\|X^{(2)}-Y^{ext}\|_{\cL^2} 
\end{align}
with 
\eqn 
	\|A\|_{\cL^2} \equiv  \sqrt{ \tr(AA^T) } \,,
\eeqn
where $A^T$ is the transpose of the matrix $A$.
Equivalently,
\begin{align}
    \mathcal{C}[W_{j}, b_{j}] &= \sqrt{\frac1N\sum_{j=1}^Q \sum_{i=1}^{N_j}
	|W_2\sigma(W_1 x_{0,j,i}+b_1)+b_2-y_j|^2_{\R^Q} }.
\end{align}

The cost is a function of the weights and biases, while the training inputs in $X_0$
are considered to be a fixed set of parameters. 
Training this shallow network amounts to finding the optimizers $W_j^{**},b_j^{**}$ for
the minimization problem
\eqn\label{eq-cC-min-1-0}
	\cC[W_j^{**},b_j^{**}] = \min_{W_j,b_j}\cC[W_j,b_j] \,.
\eeqn

\section{Statement of Main Results}
\label{sec-MainRes-1}

In this section, we present the main results of this paper, an explicit upper bound on the
cost function in Theorem \ref{thm-cC-uppbd-1} which leads to the construction of an
optimized constructively trained shallow network, and its natural geometric interpretation
in Theorem \ref{thm-DL-geometry-1}. In Theorem \ref{thm-cC-uppbd-2}, we construct an
explicit degenerate local minimum of the cost function, for the special case of input and
output space having equal dimension, $M=Q$. In Theorem \ref{thm-CostN-truc-1-0}, we
address the effects of truncation of the training input matrix due to the activation
function, in the case $Q=M$.

All bounds derived here make use of the following quantity,
\eqn 
	\delta_P := \sup_{i,j}\left|\Pen[\overline{X_0^{red}}]P\Delta x_{0,j,i}\right| \,,
\eeqn 
which measures the relative size between $\overline{X_0^{red}}$ and $P\Delta X_0$, as
$\left|\Pen[\overline{X_0^{red}}]P\Delta x_{0,j,i}\right|$ scales like the noise-to-signal
ratio  $\frac{|\Delta x|}{|x|}$ of training inputs.
Note that it is scaling invariant
under $X_0\rightarrow\lambda X_0$.  

\subsection{Upper bound on minimum of cost function for \texorpdfstring{$M > Q$}{M > Q}}

In Theorem \ref{thm-cC-uppbd-1}, we prove an upper bound on the minimum of the cost
function with the following construction.

Let $R\in O(M)$ denote an orthogonal matrix that diagonalizes $P$, defined in
\eqref{orthoprojector}. That is, the
orthoprojectors 
\eqn\label{eq-PR-def-0-1}
	P_R:=RPR^T
	\;\;\;{\rm and} \;
	P^\perp_R:=RP^\perp R^T
\eeqn
are diagonal in the given coordinate system; this is important for compatibility with the
fact that $\sigma$ acts component-wise. Namely, $R$ rotates the input data in such a way
that $\ran(P)$ is made to align with the coordinate axes; this decouples the action of
$\sigma$ on the rotated $\ran(P)$ from its action on the rotated $\ran(P^\perp)$. That is,
$\sigma(v)=\sigma(P_R v)+\sigma(P^\perp_R v)$ for all $v\in\R^M$.

%The choice of $R$ is not unique; it can be composed with any $R'\in O(M)$ which leaves the ranges of $P_R$ and $P_R^\perp$ invariant, and can be chosen such that
%\eqn 
%	R \overline{x} \;{\rm is \; parallel \; to \;} P_R u_M \;\in\R^M
%\eeqn
%where 
Letting
\eqn\label{eq-uM-def-1-0-0}	
	u_M:=(1,1,\cdots,1)^T \in\R^M
\eeqn 
%In particular, for 
we may choose $\beta_1\geq0$ sufficiently large ($\beta_1\geq 2\max_{j,i}|x_{0,j,i}|$ is
sufficient), so that the projected, translated, and rotated training input vectors $R P
x_{0,j,i}+\beta_1 u_M$ are component-wise non-negative for all $i=1,\dots,N_j$, and
$j=1,\dots,Q$.

Then, we construct an upper bound on the cost function by use of the following weights and
biases.
We choose
\eqn 
	W_1^* = R
\eeqn
and  $b_1^*=P_R b_1^*+P_R^\perp b_1^*$ where
\eqn
	P_R b_1^* =\beta_1 P_R u_Q %R \overline{x}
	\;\;\;,\;\;\;
	P^\perp_R b_1^*  = - \delta P^\perp_R u_M \,,
\eeqn
%for an arbitrary constant $\mu\geq 1$, and
with $\beta_1\geq 0$ large enough to ensure component-wise non-negativity of $RPX_0+P_R
B_1^*$, where $B_1^*=b_1^* u_N^T$, so that 
\eqn\label{eq-sig-PR-split-1-0-0}
	\sigma(RP X_0 + P_R B_1^*) = RP X_0 + P_R B_1^* \,.
\eeqn 
On the other hand, the $M-Q$ rows of $RP^\perp X_0+P_R^\perp B_1^*$ are eliminated by way of
\eqn
	\sigma(RP^\perp X_0+P_R^\perp B_1^*) =0 \,.
\eeqn
We thus obtain a reduction of the dimension of the input space from $M$ to $Q$.
  
Passing to the output layer, we require $W_2^*$ to solve
\eqn
	W_2^* R\overline{X_0^{red}} = Y \,,
\eeqn
which yields 
\eqn 
	W_2^* =  Y \; \Pen[\overline{X_0^{red}}] P R^T \,.
\eeqn
% Here,
% \eqn 
% 	\Pen[\overline{X_0^{red}}]
% 	:= ((\overline{X_0^{red}})^T\overline{X_0^{red}})^{-1}
% 	(\overline{X_0^{red}})^T
% \eeqn
% denotes the Penrose inverse of $\overline{X_0^{red}}$, which satisfies 
% \eqn	
% 	\Pen[\overline{X_0^{red}}]\overline{X_0^{red}}=\1_{Q\times Q}
% 	\;\;\;,\;\;
% 	\overline{X_0^{red}}\Pen[\overline{X_0^{red}}]=P \,.
% \eeqn 
Finally, we find that
\eqn
	b_2^*= - W_2^* P_R b_1^* \,,
\eeqn
that is, it reverts the translation by $P_R b_1^*$ in the previous layer.

\begin{theorem}\label{thm-cC-uppbd-1}
%Let
%\eqn\label{eq-tildW-def-1} 
%	\widetilde W_2 := Y \; \Pen[\overline{X_0^{red}}] \in \R^{Q\times M} \,,
%\eeqn 
%where the orthogonal matrix $R\in O(M)$ solves the diagonalization problem in \eqref{eq-P-diag-1}, below.
Let $Q\leq M \leq QM$. 
Assume that $R\in O(M)$ diagonalizes $P,P^\perp$, and let $\beta_1\geq 2\max_{j,i}|x_{0,j,i}|$. 
% are such that the condition \eqref{eq-angle-cond-1} is satisfied.

Let $\cC[W_i^*,b_i^*]$ be the cost function evaluated for the trained shallow network defined by the following weights and biases,
\eqn 
	W_1^*=R \,,
\eeqn
and
\eqn\label{eq-tildW2-def-1-1}
	W_2^* = \widetilde W_2 R^T
	\;\;\;{\rm where} \;
	\widetilde W_2 = Y \; \Pen[\overline{X_0^{red}}] P \,.
\eeqn 
Moreover, $b_1^*=P_R b_1^*+P_R^\perp b_1^*$ with
\eqn 
	&&
	P_R b_1^*= \beta_1 u_M % R\overline{x}  
	\;\;,\; 
	P_R^\perp b_1^* = - \delta P_R^\perp u_M \;\; \in \R^M \,,
\eeqn
for %an arbitrary $\mu>1$, 
$u_M\in\R^M$ as in \eqref{eq-uM-def-1-0-0}, and
\eqn 
	b_2^* = -  W_2^* P_R b_1^* \;\; \in\R^Q \,.
\eeqn 
Then, the minimum of the cost function satisfies the upper bound
\eqn\label{eq-cC-uppbd-thm-1}
	\min_{W_j,b_j}\cC[W_j,b_j] \leq  \cC[W_i^*,b_i^*] \leq 
	\frac1{\sqrt N}\|Y \; \Pen[\overline{X_0^{red}}] P\Delta X_0  \|_{\cL^2} 
%	\|\widetilde W_2 P\Delta X_0 -Av(\widetilde W_2P\Delta X_0 ) \|_{\cL^2} 
	\,,
\eeqn 
%where the row-wise average $Av(A)$ of a matrix $A=[a_{ij}]\in\R^{M\times N}$ is given by 
%\eqn\label{eq-Av-def-1}
%	[Av(A) ]_{ij} := 
%	\frac1N\sum_{j=1}^N a_{ij} 
%\eeqn 
%for every $i\in\{1,\dots,Q\}$.
%
which implies
\eqn\label{eq-cC-average-bd-1}
	 \min_{W_j,b_j}\cC[W_j,b_j] \; \leq \; 
	 \|Y\|_{op} \;\delta_P \;
	%\sqrt{N} 
\eeqn 
where $N=\sum_{\ell=1}^Q N_\ell$. 
 
\end{theorem}

A geometrically intuitive way of thinking about this construction is that $W_1^*=R$
orients the training input data with respect to the given coordinate system, in order to
align it with the component-wise action of $\sigma$. This allows for a maximal rank
reduction via $\sigma$, whereby the maximal possible amount of insignificant information
is eliminated: $P_Rb_1^*$ pulls the significant information (in the range of $P_R$) out of
the kernel of $\sigma$, while $P_R^\perp b_1^*$ pushes the insignificant information (in
the range of $P_R^\perp$) into the kernel of $\sigma$ whereby the latter is eliminated.
Subsequently, $b_2^*$ places the significant information back into its original position,
and $W_2^*$ matches it to the output matrix $Y$ in the sense of least squares, with
respect to the $\cL^2$-norm.

We observe that, for any $\lambda>0$, the rescaling of training inputs
\eqn
	x_{0,j,i}\rightarrow \lambda x_{0,j,i}
	\;\;\forall j=1,\dots,Q
	\;\;,\;i=1,\dots,N_j \,,
\eeqn 
induces 
\eqn
	\Pen[\overline{X_0^{red}}] \rightarrow \lambda^{-1}\Pen[\overline{X_0^{red}}] 
	\;\; \;,\;
	\Delta X_0 \rightarrow \lambda \Delta X_0 \,.
\eeqn 
Therefore,  the upper bound in \eqref{eq-cC-uppbd-thm-1} is scaling invariant, as it only
depends on the signal-to-noise ratio of training input data, which is controlled by
$\delta_P$.

\subsection{Exact degenerate local minimum in the case \texorpdfstring{$M=Q$}{M=Q}}

We explicitly determine an exact local degenerate minimum of the cost function in the case
$Q=M$ where the input and output spaces have the same dimension; here, we show that the
upper bound obtained above differs from the sharp value by a relative error of order
$O(\delta_P^2)$. We note that in this situation, the projector defined in
\eqref{orthoprojector} satisfies $P=\1_{Q\times Q}$ and $P^\perp=0$.
Moreover, $\overline{X_0^{red}}$ is invertible, and therefore,
$\Pen[\overline{X_0^{red}}]=(\overline{X_0^{red}})^{-1}$. 

We prove this result relative to a weighted variant of the cost function defined as follows.
Let $\cN\in\R^{N\times N}$ be the block diagonal matrix given by
\eqn
	\cN:=\diag(N_j \1_{N_j\times N_j} \, | \, j=1,\dots,Q) \,.
\eeqn 
We introduce the inner product on $\R^{Q\times N}$
\eqn	
	(A,B)_{\cL^2_\cN}:=\tr(A\cN^{-1}B^T)
\eeqn
and 
\eqn	
	\|A\|_{\cL^2_\cN}:=\sqrt{(A,A)_{\cL^2_\cN}} \,.
\eeqn 
We define the weighted cost function
\eqn\label{eq-CostN-def-1-0-0}
	\CostN[W_i,b_i] := \|X^{(2)}-Y^{ext}\|_{\cL^2_\cN} \,.
\eeqn 
This is equivalent to 
\eqn\label{eq-CostN-def-1}
	\CostN[W_i,b_i]:= \sqrt{\sum_{j=1}^Q \frac{1}{N_j}\sum_{i=1}^{N_j}
	|W_2\sigma(W_1 x_{0,j,i}+b_1)+b_2-y_j|^2_{\R^Q} } \,.
\eeqn 
The weights $\frac 1{N_j}$ ensure that the contributions to the cost function belonging to inputs $y_j$ do not depend on their sample sizes $N_j$. We note that for uniform sample sizes, where $\frac{N_j}{N}=\frac1Q$ $\forall j$, we have that $\CostN[W_i,b_i]=\sqrt{Q }\cC[W_i,b_i]$.

\begin{theorem}
\label{thm-cC-uppbd-2}
Assume $M=Q<MQ$.  
Then, if $\rank(\overline{X_0})=Q$, let
\eqn\label{eq-cP-def-1-0-0}
	\cP := \cN^{-1}X_0^T(X_0 \cN^{-1} X_0^T)^{-1} X_0 
	\;\;
	\in\R^{N\times N} \,.
\eeqn 
Its transpose $\cP^T$ is an orthonormal projector in $\R^N$ with respect to the inner product $\langle u,v\rangle_{\cN}:=(u,\cN^{-1} v)$. In particular,  $(A\cP, B)_{\cL^2_\cN}=(A,B\cP)_{\cL^2_\cN}$ for all $A,B\in\R^{Q\times N}$. 
Letting $\cP^\perp=\1_{N\times N}-\cP$, the weighted cost function satisfies the upper bound
\eqn\label{eq-Thm3.2-CostN-1-0}
	\min_{W_j,b_j}\CostN[W_j,b_j]&\leq&\CostN[W_i^*,b_i^*]
	\nonumber\\
	&=&
	\|Y^{ext}\cP^{\perp}\|_{\cL^2_{\cN}} 
	\nonumber\\
	&=&
	\big\|
 	Y|\Delta_2^{rel}|^{\frac12}\big(1+\Delta_2^{rel}\big)^{-\frac12}   
 	\big\|_{\cL^2}
	\nonumber\\
	&\leq&(1-C_0\delta_P^2) \; \|Y \; (\overline{X_0^{red}})^{-1}\Delta X_0  \|_{\cL^2_{\cN}} 
	\,,
\eeqn 
for a constant $C_0\geq0$, and where
\eqn
	\Delta_2^{rel} := 
	\Delta_1^{rel} \cN^{-1} (\Delta_1^{rel})^{T} 
	\;\;,\;{\rm with}\;
	\Delta_1^{rel} :=  
	(\overline{X_0^{red}})^{-1} 
	\Delta X_0\,.
\eeqn 
The weights and biases realizing the upper bound are given by
\eqn 
	W_1^*=\1_{Q\times Q} 
%	\;\;\;
%	{\rm such\;that\;}
%	\;\;
%	R\overline{x}=|\overline{x}|\frac{u_Q}{|u_Q|}\,,
\eeqn
and
\eqn
	W_2^* 
	= Y(\overline{X_0^{red}})^T(X_0 \cN^{-1} X_0^T)^{-1} \in\R^{Q\times Q} \,.
\eeqn 	
Moreover,
\eqn\label{eq-Thm3.2-b1b2-1}
	b_1^*=\beta_1 u_Q
	%R\overline{x} 
	\;\;,\; b_2^*=-W_2^*b_1^* \,,
\eeqn 
with $u_Q\in\R^Q$ as in \eqref{eq-uM-def-1-0-0}, 
and where $\beta_1\geq 2\max_{j,i}|x_{0,j,i}|$ so that 
\eqn\label{eq-W1B1-invar-1-0-0}
	\sigma(W_1^*X_0+B_1^*)=W_1^*X_0+B_1^* \,.
\eeqn  
Notably,
\eqn 
	W_2^*=\widetilde W_2 +O(\delta_P^2) 
\eeqn
differs by $O(\delta_P^2)$ from $\widetilde W_2$  in \eqref{eq-tildW2-def-1-1} used for the upper bound \eqref{eq-cC-uppbd-thm-1}.

In particular, $\CostN[W_i^*,b_i^*]$ is a local minimum of the cost function; it is degenerate, assuming the same value for all $W_i,b_i$ such that $W_1,b_1$ satisfy the condition \eqref{eq-W1B1-invar-1-0-0}.

Moreover, $\CostN[W_i^*,b_i^*]$ is invariant under reparametrizations of the training inputs $X_0\rightarrow K X_0$, for all $K\in GL(Q)$.
\end{theorem}

The proof is given in Section \ref{sec-prf-thm-cC-uppbd-2}.

\subsection{Geometric interpretation} 
To match an arbitrary test input $x\in\R^M$ with one of the output vectors $y_j$, $j\in\{1,\dots,Q\}$, let for $x\in\R^M$,  
\eqn 
	\cC_j[x] := |W_2^*(W_1^* x + b_1^*)_+ + b_2^*-y_j| \,.
\eeqn 
Here, $(\cdot)_+$ denotes the ramp function \eqref{eq-rampfct=def-1}, acting component-wise on a vector. 

Given an arbitrary test input $x\in\R^M$, 
\eqn 
	j^*= \argmin_j \cC_j[x] 
\eeqn 
implies that $x$ matches the $j^*$-th output $y_{j^*}$.

The proof of Theorem \ref{thm-cC-uppbd-1} provides us with a particular set of weights
$W_1^*,W_2^*$ and biases $b_1^*,b_2^*$ through an explicit construction, which yield the
upper bound \eqref{eq-cC-uppbd-thm-1}.  We will refer to the shallow network trained with
this choice of $W_i^*,b_i^*$ as the {\em constructively trained shallow network}. 
% A detailed discussion is provided in Section \ref{sec-theor-DL-1}.

\begin{theorem}\label{thm-DL-geometry-1}
Assume $Q\leq M\leq QM$.
Let $W_i^*,b_i^*$, $i=1,2$, denote the weights and biases determined in Theorem \ref{thm-cC-uppbd-1}. 
% and Section \ref{sec-theor-DL-1}.
Let
\eqn\label{eq-tildW-def-1} 
	\widetilde W_2 := Y \; \Pen[\overline{X_0^{red}}] \in \R^{Q\times M} \,,
\eeqn 
and define the metric 
\eqn\label{eq-dW2-def-1}
	d_{\widetilde W_2}(x,y):=|\widetilde W_2 P(x - y)| 
	\;\;\;{\rm for}\;
	x,y\in \ran(P)
\eeqn 
on the $Q$-dimensional linear subspace $\ran(P)\subset\R^M$, where $|\cdot|$ denotes the Euclidean norm on $\R^Q$.
Then, 
\eqn
	\cC_j[x] = d_{\widetilde W_2}(Px,\overline{x_{0,j}}) \,,
\eeqn 
and matching an input $x\in\R^M$ with a target output $y_{j^*}$ via the
constructively trained shallow network is equivalent to the solution of the metric
minimization problem
\eqn 
	j^* = \argmin_{j\in\{1,\dots,Q\}} (d_{\widetilde W_2}(Px,\overline{x_{0,j}}))
\eeqn 
on the range of $P$. 
\end{theorem}

The proof is given in Section \ref{sec-prf-thm-DL-geometry-1}. The geometric interpretation of the construction is as follows.

First of all, we note that $\widetilde W_2 P = \widetilde W_2$ has full rank $Q$, therefore $x\mapsto|\widetilde W_2 P x|^2$ is a non-degenerate quadratic form on the range of $P$. Therefore, \eqref{eq-dW2-def-1} indeed defines a metric on the range of $P$.

The constructively trained shallow network thus obtained matches a test input $x\in\R^M$
with an output vector $y_{j^*}$  by splitting
\eqn 
	x = Px + P^\perp x \,,
\eeqn 
and by determining which of the average training input vectors $\overline{x_{0,j}}$, $j=1,\dots,Q$, is closest to $Px$ in the $d_{\widetilde W_2}$ metric. The trained shallow network cuts off the components $P^\perp x$.

\subsection{Dependence on truncation}
\label{ssec-truncation}

Our next result addresses the effect of truncations enacted by the activation function $\sigma$ in the case $M=Q$.
For the results in Theorem \ref{thm-cC-uppbd-2}, we assumed that $W_1,b_1$ are in the region (depending on $X_0$) such that 
\eqn\label{eq-sig-triv-1-0-0}
	\sigma(W_1 X_0+B_1) = W_1 X_0+B_1
\eeqn 
holds. Here we discuss the situation in which $\sigma$ acts nontrivially. In this case, we assume that $W_1, b_1$ are given, with $W_1\in\R^{Q\times Q}$ invertible, and observe that all matrix components of
\eqn 
	X^{(1)} = \sigma(W_1 X_0+B_1)
\eeqn 
are non-negative, as $X^{(1)}$ lies in the image of $\sigma$, and hence,  
\eqn 
	\sigma(X^{(1)} ) = X^{(1)}  \,.
\eeqn
We define the truncation map $\tau_{W_1,b_1}$ as follows. It will play a major role in our
explicit construction of local and global cost minimizers in deep neural networks in
\cite{chenewald23deep}.

\begin{definition}
Let $W_1\in GL(M)$, $b_1\in\R^M$. Then, the truncation map $\tau_{W_1,b_1}:\R^{M\times
N}\rightarrow \R^{M\times N}$ is defined by 
\eqn 
	\tau_{W_1,b_1}(X_0) &:=& 
	W_1^{-1} ( \sigma(W_1 X_ 0 + B_1) - B_1)
	\nonumber\\
	&=&
	W_1^{-1} ( X^{(1)} - B_1) \,.
\eeqn 
That is, $\tau_{W_1,b_1} = a_{W_1,b_1}^{-1} \circ\sigma\circ a_{W_1,b_1} $ under the affine map $a_{W_1,b_1} :X_0\mapsto W_1X_0+B_1$.

We say that $\tau_{W_1,b_1}$ is rank preserving (with respect to $X_0$) if both
\eqn
	\rank(\tau_{W_1,b_1}(X_0))&=&\rank(X_0)
	\nonumber\\
	\rank(\overline{\tau_{W_1,b_1}(X_0)})&=&\rank(\overline{X_0})
\eeqn 
hold, and that it is rank reducing otherwise.	
\end{definition}

Then, we verify that
\eqn\label{eq-sig-invar-3-0-0}
	\sigma(W_1 \tau_{W_1,b_1}(X_0)  + B_1) &=& \sigma(X^{(1)}) = X^{(1)} 
	\nonumber\\
	&=& W_1 \tau_{W_1,b_1}(X_0)  + B_1 \,.
\eeqn 
This means that while the condition \eqref{eq-sig-triv-1-0-0} does not hold for $X_0$, it
does hold for $\tau_{W_1,b_1}(X_0)$.

Accordingly, we can define the matrices 
\eqn 
	\overline{\tau_{W_1,b_1}(X_0) }
	\;\;,\;
	\overline{(\tau_{W_1,b_1}(X_0) )^{red}}
	\;\;,\;
	\Delta (\tau_{W_1,b_1}(X_0) )
\eeqn 
in analogy to $\overline{X_0}, \overline{X_0^{red}}, \Delta X_0$.

If no rank reduction is induced by the truncation, $\overline{(\tau_{W_1,b_1}(X_0) )^{red}}$ is invertible, and we obtain the following theorem.

\begin{theorem}
\label{thm-CostN-truc-1-0}
Let $Q=M$, and assume that the truncation map $\tau_{W_1,b_1}(X_0) $ is rank preserving. Then, for any fixed $(W_1,b_1)\in GL(Q)\times \R^Q$,
\eqn\label{eq-minW2b2-CostN-1-0} 
	\min_{W_2,b_2}\CostN[W_i,b_i] = 
	\|Y^{ext} \cP^\perp_{\tau_{W_1,b_1}(X_0)} \|_{\cL^2_{\cN}}
\eeqn 
where the projector $\cP_{\tau_{W_1,b_1}(X_0)}$ is obtained from \eqref{eq-cP-def-1-0-0} by substituting the truncated training inputs $\tau_{W_1,b_1}(X_0)$ for $X_0$. 
In particular, one explicitly has
\eqn\label{eq-minW2b2-CostN-2-0} 
	\min_{W_2,b_2}\CostN[W_i,b_i]
	&=&
	\big\|
 	Y|\Delta_2^{rel,tr}|^{\frac12}\big(1+\Delta_2^{rel,tr}\big)^{-\frac12}   
 	\big\|_{\cL^2}
	\nonumber\\
	&\leq& 
	(1-C_0\delta_{P,tr}^2) \|Y  \Delta_1^{rel,tr} \|_{\cL^2_{\cN}}
\eeqn 
where
\eqn
	\Delta_2^{rel,tr} &:=& 
	\Delta_1^{rel,tr} \cN^{-1} (\Delta_1^{rel,tr})^{T} 
	\;\;,\;{\rm with}\;
	\nonumber\\
	\Delta_1^{rel,tr} &:=&  
	(\overline{(\tau_{W_1,b_1}(X_0) )^{red}})^{-1}
	\Delta (\tau_{W_1,b_1}(x_{0,j,i}) )\,,
\eeqn 
in analogy to \eqref{eq-Thm3.2-CostN-1-0}, and for some $C_0\geq0$, where
\eqn 
	\delta_{P,tr} := \sup_{j,i}\left|(\overline{(\tau_{W_1,b_1}(X_0) )^{red}})^{-1}
	\Delta (\tau_{W_1,b_1}(x_{0,j,i}) )\right|
\eeqn
measures the signal-to-noise ratio of the truncated training input data.
\end{theorem}

\prf
Due to \eqref{eq-sig-invar-3-0-0}, we may follow the proof of Theorem \ref{thm-cC-uppbd-2} verbatim, and we find that for the given fixed choice of $W_1,b_1$, minimization in $W_2,b_2$ yields \eqref{eq-minW2b2-CostN-1-0} and \eqref{eq-minW2b2-CostN-2-0}.
\endprf 

In general, \eqref{eq-minW2b2-CostN-2-0} does not constitute a stationary solution with respect to $W_1,b_1$, as the right hand sides of \eqref{eq-minW2b2-CostN-1-0}, \eqref{eq-minW2b2-CostN-2-0}  generically depend nontrivially on $W_1,b_1$. It thus remains to determine the infimum of \eqref{eq-minW2b2-CostN-1-0}, \eqref{eq-minW2b2-CostN-2-0} with respect to  $W_1,b_1$, in order to find a candidate for the global minimum. The latter, however, might depend sensitively on the detailed properties of $\Delta X_0$, which are random.

On the other hand, in the situation addressed in Theorem \ref{thm-cC-uppbd-2}, $W_1,b_1$ are contained in the parameter region $\cF_{X_0}$ (depending on $X_0$) defined by the fixed point relation for the truncation map (equivalent to \eqref{eq-sig-triv-1-0-0}),
\eqn
	\cF_{X_0}:=\{(W_1,b_1)\in GL(Q)\times \R^Q \; | \; \tau_{W_1,b_1}(X_0) = X_0 \}\,.
\eeqn
As a result, we indeed obtained a degenerate stationary solution with respect to $W_1,b_1$, 
\eqn\label{eq-minW2b2-CostN-deg-1-0} 
	\min_{(W_1,b_1)\in \cF_{X_0};W_2,b_2}\CostN[W_i,b_i] 
	= \|Y^{ext}\cP^{\perp}\|_{\cL^2_{\cN}} 
\eeqn 
in \eqref{eq-Thm3.2-CostN-1-0}, with $\cP$ given in \eqref{eq-cP-def-1-0-0}.  
Hence, $\cF_{X_0}$ parametrizes an invariant manifold of equilibria of the gradient descent flow, which corresponds to a level surface of the cost function. 

\begin{remark}
The global minimum of the cost function has either the form \eqref{eq-minW2b2-CostN-1-0} for an optimizer $(W_1^{**},b_1^{**})$ (with a rank preserving truncation), or \eqref{eq-minW2b2-CostN-deg-1-0}. This is because any other scenario employs a rank reducing truncation. Since $Y^{ext}$ has rank $Q$, minimization of \eqref{eq-CostN-def-1-0-0} under the constraint $\rank(\tau_{W_1,b_1}(X_0))<Q$ cannot yield a global minimum.

The key point in minimizing \eqref{eq-minW2b2-CostN-1-0} with respect to $(W_1,b_1)$ is to determine a rank preserving truncation which minimizes the signal to noise ratio of the truncated training inputs. Whether for a local or global minimum of the form \eqref{eq-minW2b2-CostN-1-0} or \eqref{eq-minW2b2-CostN-deg-1-0}, respectively, a shallow network trained with the corresponding weights and biases $W_i,b_i$, $i=1,2$, will be able to match inputs $x\in\R^Q$ to outputs $y_j$, through a straightforward generalization of Theorem \ref{thm-DL-geometry-1}. 
\end{remark}

 We expect these considerations to carry over to the general case $Q<M$, but leave a detailed discussion to future work. 

\section{Experiments} \label{experiments}

\begin{figure}[h]
    \includegraphics[width=0.49\textwidth]{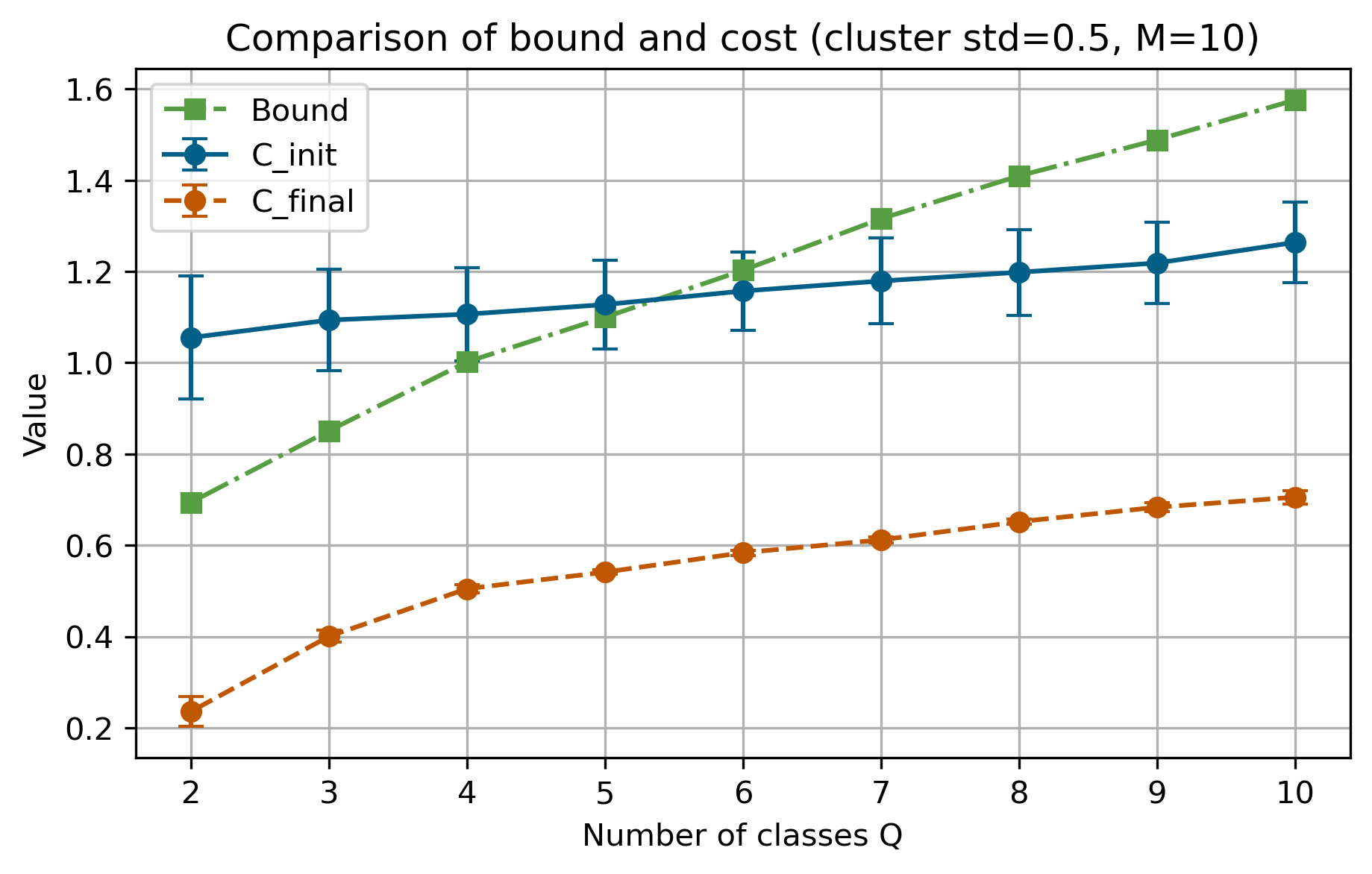}
    \includegraphics[width=0.49\textwidth]{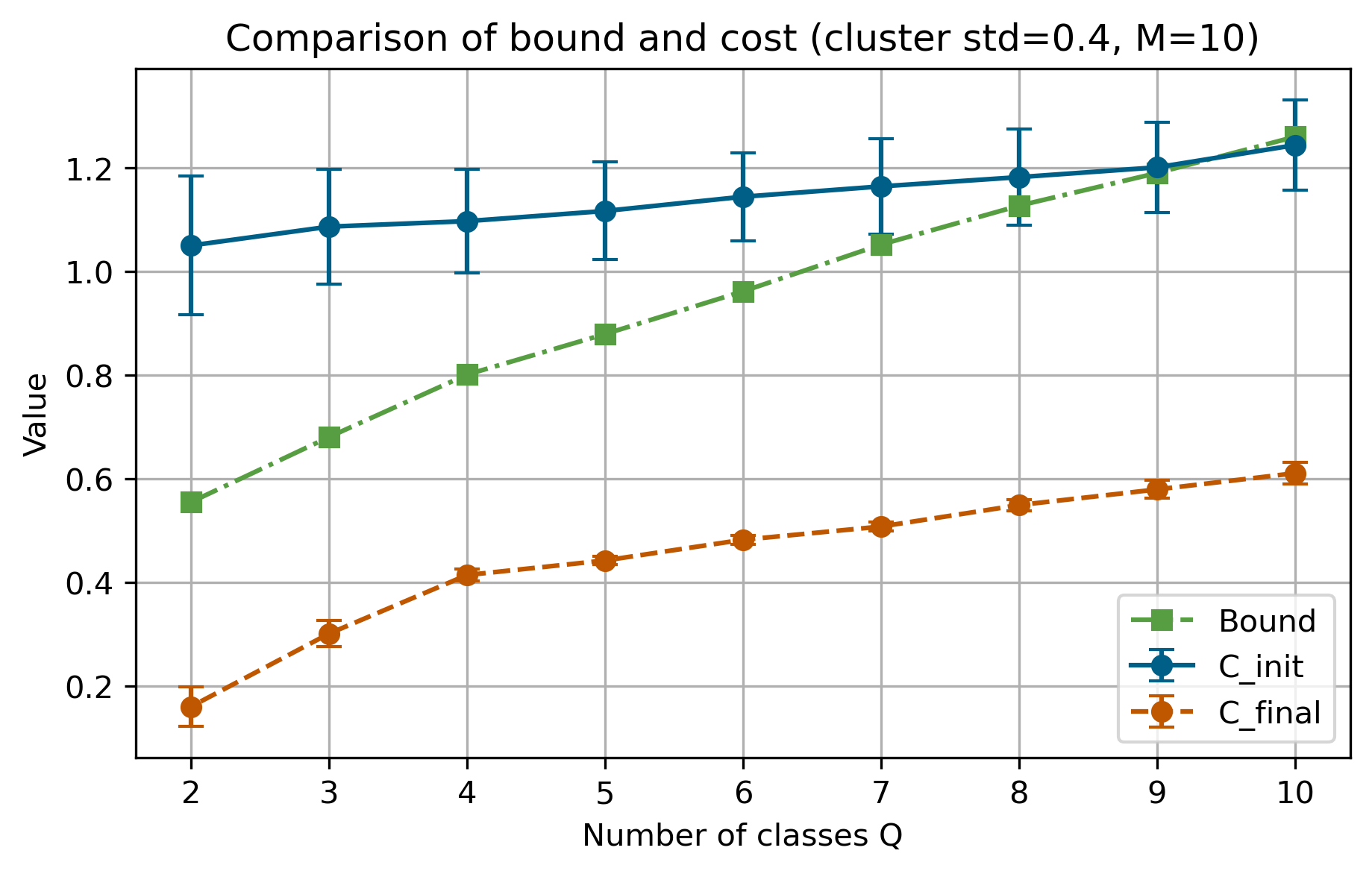}
    \includegraphics[width=0.49\textwidth]{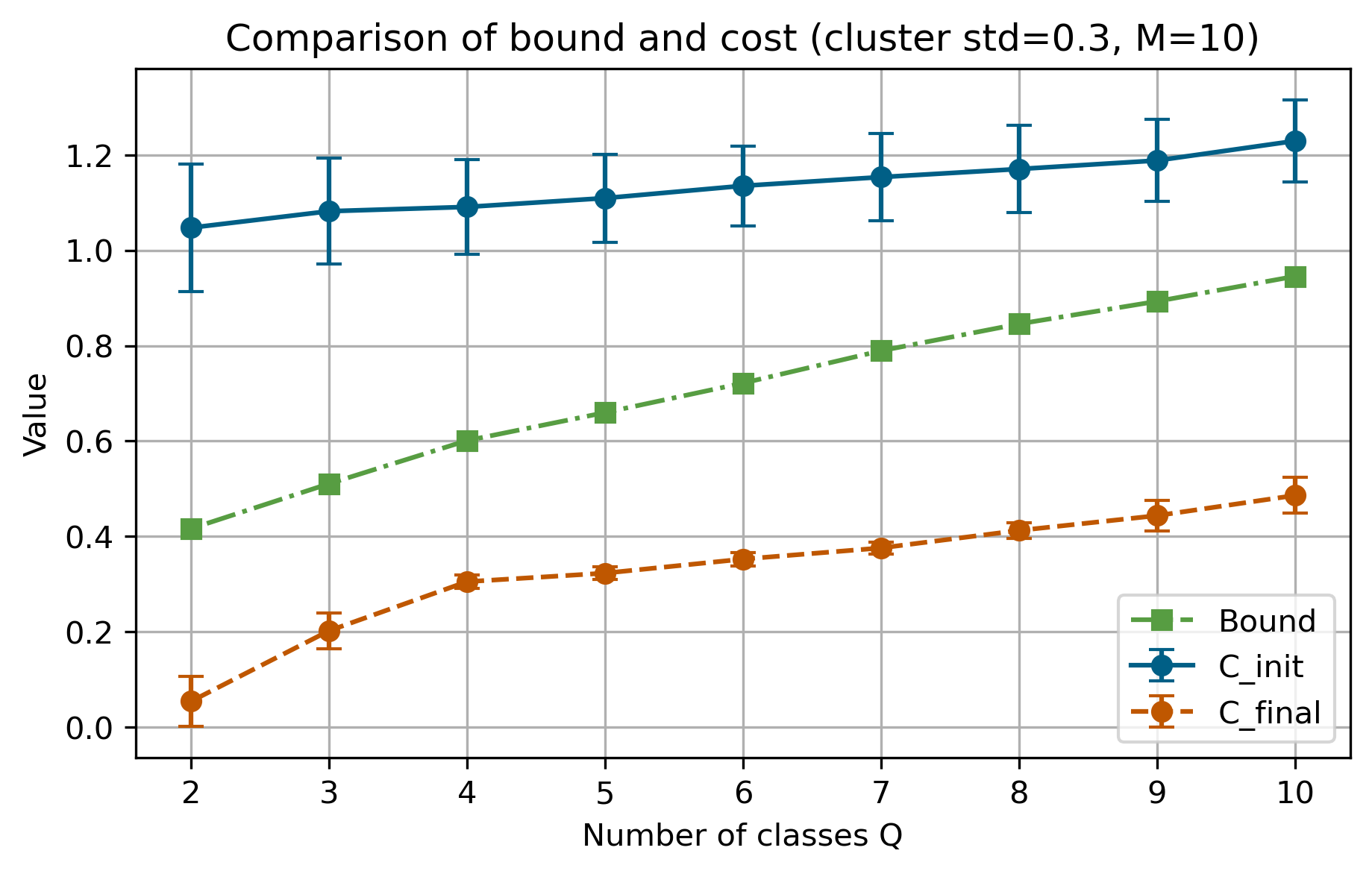}
    \includegraphics[width=0.49\textwidth]{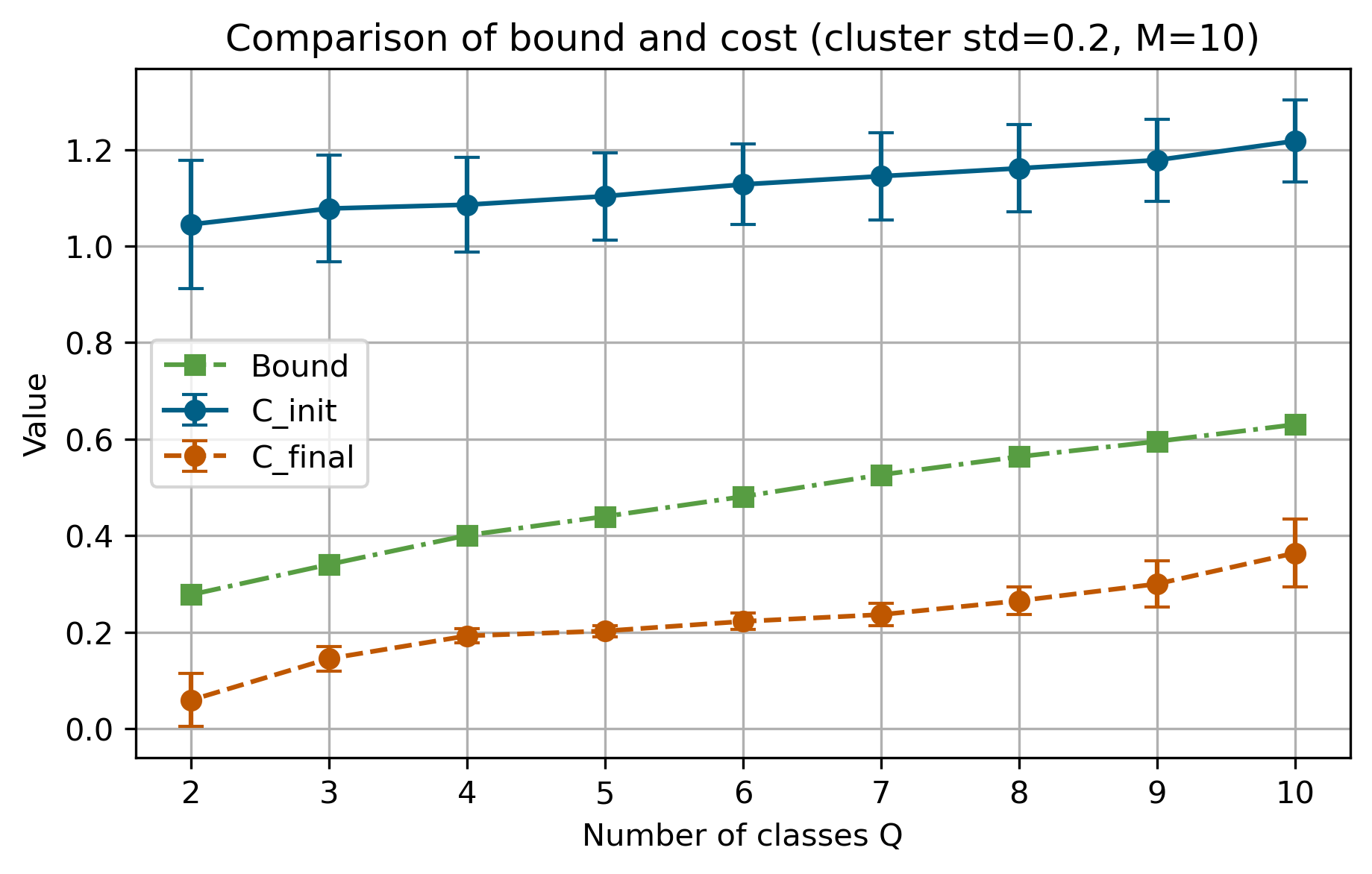}
    \caption{(Average of) initial ($C_{init}$) and final ($C_{final}$) cost of randomly
        initialized neural network(s) trained to classify Gaussian mixture data with $Q$
        classes, for different (fixed in each plot) cluster standard deviations,
    plotted against the bound \eqref{boundexp} computed for each data set.} 
    \label{plots}
\end{figure}

\begin{figure}[h]
    \centering
    \includegraphics[width=0.9\textwidth]{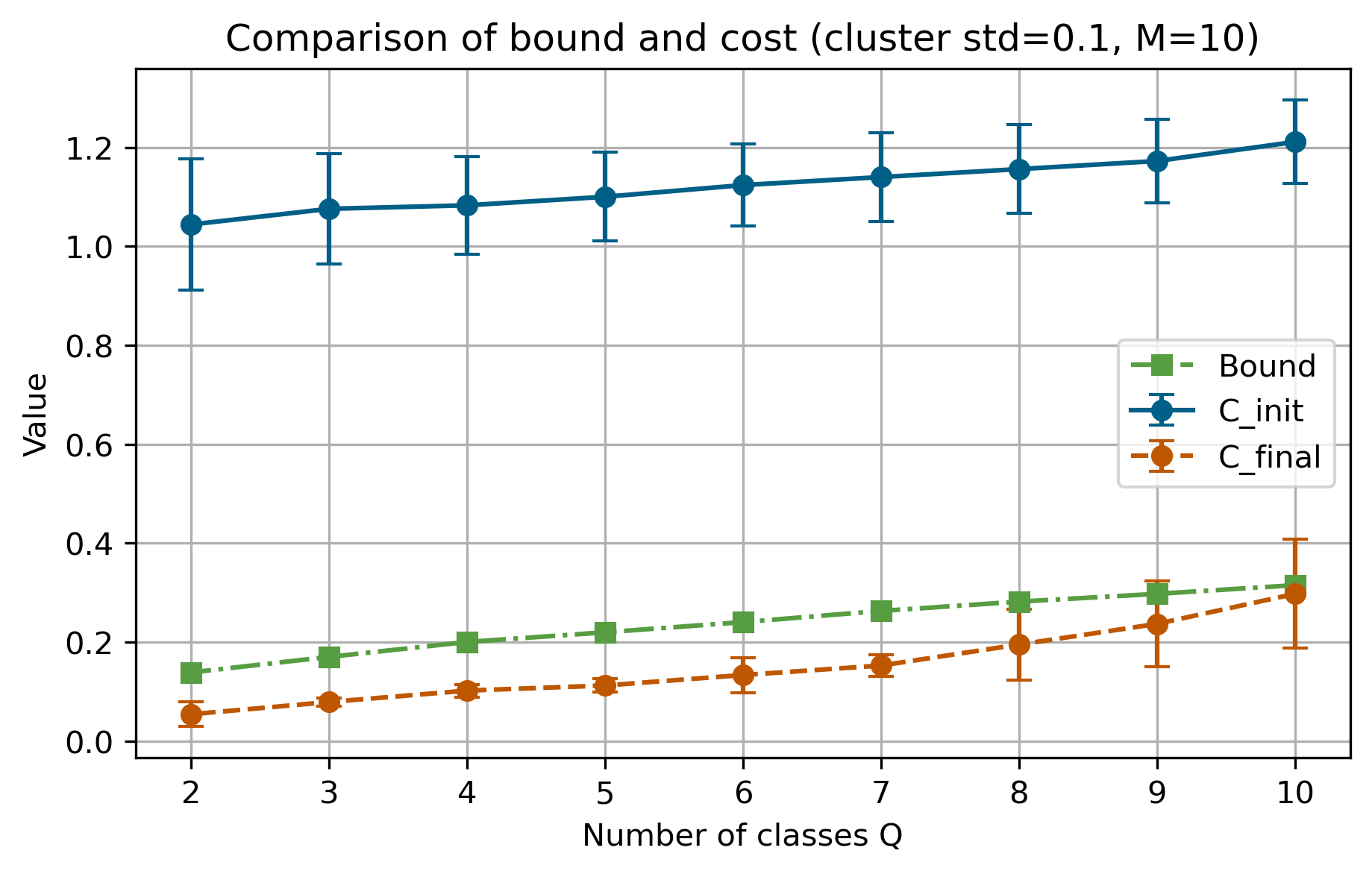}
    \caption{(Average of) initial ($C_{init}$) and final ($C_{final}$) cost of randomly
        initialized neural network(s) trained to classify Gaussian mixture data with $Q$
    classes, plotted against the bound \eqref{boundexp} computed for each data set.}
    \label{plot1}
\end{figure}

To assess the validity of our theoretical results in practical training scenarios, we
generated synthetic data using a Gaussian mixture model in $\mathbb{R}^{M}$ with $Q$
classes, each having randomly distributed means and varying standard deviations. For each
dataset, constructed with a fixed standard deviation and $Q$ classes, we initialized and
trained a ReLU neural network with architecture $(M, M, Q)$, as defined in
\eqref{neuralnet1}-\eqref{neuralnet2}, to classify the data according to the cost function
\eqref{eq-cC-def-1}. Each configuration was trained 100 times. Figures \ref{plots} and
\ref{plot1} display the average final cost across
runs, the corresponding average initial cost, and the bound \eqref{eq-cC-uppbd-thm-1} from
Theorem \ref{thm-cC-uppbd-1}, which we repeat here:
\begin{align}
    \label{boundexp} 
    \text{Bound} = \frac{1}{\sqrt{N}}\norm{\left(\overline{X_0^{red}}\right)^{+} \Delta
    X_0}_{\mathcal{L}^{2}}.
\end{align}
Note that we taking the label vectors to be the
canonical basis vectors in $\mathbb{R}^{Q}$, so that
$Y$ is the identity.

Figure \ref{plots} shows that as the variance of the clusters in the Gaussian mixture
decreases, the theoretical bound increasingly aligns with the average final cost achieved
during training.
In particular, in Figure \ref{plot1}, corresponding to the smallest variance, 
the bound closely tracks the final training loss and, for some runs, is even smaller than
the attained cost.

We summarize the experiment parameters, as well as the hyperparameters used in the
training of the neural networks via stochastic gradient descent, in Table
\ref{tab:hyperparams}. Code is available at \url{https://github.com/patriciaewald/bounds}.

\begin{table}[htbp]
    \centering
    \label{tab:hyperparams}
    \begin{tabular}{lcc}
        \hline
        Hyperparameter & Notation & Value \\
        \hline
        Input dimension & $M$ & $10$ \\
        Samples per class & $N_{j}$ & $100$ \\
        Number of classes & $Q$ & $\{10,\cdots, 2\}$ \\
        \vspace{1ex}
        Cluster standard deviation & std & $\{0.5, \cdots, 0.1\}$ \\
        Random initializations &  & $100$ \\
        Maximum number of epochs &  & $1000$ \\
        Learning rate &  & $0.05$ \\
        Momentum &  & $0.9$ \\
        Plateau tolerance &  & $10^{-4}$ \\
        Plateau patience (epochs) &  & $30$ \\
        \hline
    \end{tabular}
    \vspace{1em}
    \caption{Hyperparameters used in the Gaussian mixture and network training
    experiments.}
\end{table}

\section{Proof of Theorem \ref{thm-cC-uppbd-1}}
\label{sec-prf-thm-cC-uppbd-1}

Let $R\in O(M)$ diagonalize $P,P^\perp$ (which are symmetric),  
\eqn\label{eq-P-diag-1}
	P_R:=RPR^T
	\;\;\;,\;\;\;
	P^\perp_R:=RP^\perp R^T \,,
\eeqn
where $P_R$ has $Q$ diagonal entries of $1$, and all other components are $0$, while $P^\perp_R$ has $M-Q$ diagonal entries of $1$, and all other components are $0$.
This does not characterize $R$ uniquely, as we may compose $R$ with an arbitrary $ R'\in O(M)$ that leaves the ranges of $P,P^\perp$ invariant. 
Diagonality of $P_R,P_R^\perp$ implies that the activation function, which acts component-wise as the ramp function \eqref{eq-rampfct=def-1}, satisfies
\eqn
	\sigma(P_R v + P^\perp_R v) = \sigma(P_R v) + \sigma(P^\perp_R v)
\eeqn 
for any $v\in\R^M$, and where of course, $P_R+P_R^\perp=\1_{M\times M}$.

We write
\eqn
	b_1=P_R b_1 + P_R^\perp b_1
\eeqn
where we choose  
\eqn\label{eq-Prb1-choice-1}
	P_R b_1 = \beta_1 u_Q \,, %R\overline{x}
\eeqn 
and we pick
\eqn 
	\beta_1\geq 2\rho
	\;\;\;,\;\;
	\rho:=\max_{j,i}|x_{0,j,i}|\,.
\eeqn 
%(using $P_RRP\overline{x}=RP\overline{x}=R\overline{x}$) where $\beta_1$ is large enough that the pairwise angles between vectors $RP x_{0,j,i}+P_R b_1\in\ran(P_R)$ satisfy
%\eqn\label{eq-angle-cond-1}
%	\lefteqn{
%	\angle(RP x_{0,j,i}+P_R b_1,RP x_{0,j',i'}+P_R b_1) < \frac\pi2-\eta 
%	\;\;,\;
%	}
%	\nonumber\\
%	&&\forall
%	\;j,j'=1,\dots,Q 
%	\;\;,\;
%	\forall\;
%	i=1,\dots,N_j 
%	\;\;,\;
%	\forall\;
%	i'=1,\dots,N_{j'} 
%\eeqn 
%for some $\eta>\delta$.
%This means that all vectors $RP x_{0,j,i}+P_R b_1$ lie in a cone of opening angle $<\frac\pi2-\eta$ in the range of $P_R$.
%We may therefore compose $R$ with a rotation $R'\in O(M)$ which leaves the ranges of $P_R,P_R^\perp$ invariant, such that all vectors $R'(RP x_{0,j,i}+P_R b_1)$ lie in the positive quadrant of $\ran(P_R)$. For example, this can be achieved by choosing $R'$ such that $R'P_R b_1$ is aligned with the diagonal in $\ran(P_R)$, that is, parallel to $P_R (1,1\dots,1)^T\in\R^M$. This implies that the vector components of all $R'(RP x_{0,j,i}+P_R b_1)$ are non-negative. Hence,
This ensures that $RP x_{0,j,i}+P_R b_1 \in B_\rho(\beta_1 u_Q)\subset P_R\R_+^M$ so that
\eqn\label{eq-PX0-positivity-1}
	\sigma(RP x_{0,j,i}+P_R b_1) = RP x_{0,j,i}+P_R b_1
	\;\;\;,\;\;
	\;j=1,\dots,Q 
	\;\;,\;
	\forall\;
	i=1,\dots,N_j 
\eeqn 
To construct the upper bound, we choose
\eqn\label{eq-W1-QR-1}
	W_1 = R \,.
\eeqn 
Then, 
\eqn 
	X^{(1)}&=&\sigma(W_1 X_0 + B_1) 
	\nonumber\\
	&=& \sigma((P_R+P_R^\perp)R X_0 +(P_R+P_R^\perp)B_1)
	\nonumber\\
	&=& \sigma(P_R R  X_0+P_R B_1)+ \sigma(P^\perp_R R X_0+P^\perp_R B_1)
	\nonumber\\
	&=&\sigma(R P  X_0 +P_RB_1) + \sigma( R P^\perp \Delta X_0+P^\perp_R B_1)\,.
\eeqn 
This is because by construction,
\eqn 
	P \overline{X_0} = \overline{X_0}
	\;\;\;,\;\;
	P^\perp \overline{X_0} = 0 \,.
\eeqn 
Moreover, we used 
\eqn
	P_R R =RP 
	\;\;\;,\;\;
	P^\perp_R R =   R P^\perp \,,
\eeqn 
from \eqref{eq-P-diag-1}.

Going to the next layer, we have
\eqn 
	X^{(2)} = W_2 X^{(1)} + B_2 \,.
\eeqn 
Our goal is to minimize the expression on the right hand side of
\eqn\label{eq-cC-upperbd-1}
	\sqrt{N}\cC[W_j,b_j] &\leq&  
	\|W_2\sigma(R P  X_0 +P_RB_1) 
	+B_2-Y^{ext}\|_{\cL^2}
	\nonumber\\
	&&\hspace{1cm}
	+\|W_2\sigma(R P^\perp \Delta X_0+P^\perp_R B_1)\|_{\cL^2}\,.
\eeqn
We note that because $\rank(Y^{ext})=Q$, and
\eqn 
	\rank(\sigma(\underbrace{R P  X_0}_{=P_RRPX_0} +P_RB_1) ) \leq Q \,,
\eeqn
minimization of the first term on the r.h.s. of \eqref{eq-cC-upperbd-1} requires that the latter has maximal rank $Q$. Due to our choice of $R$ and $P_R b_1$ which imply that \eqref{eq-PX0-positivity-1} holds, we find that
\eqn\label{eq-PRB1-req-1} 
	\sigma(R P  X_0 +P_RB_1)
	&=&
	R P  X_0 +P_RB_1
	\nonumber\\
	&=&
	P_R R ( \overline{X_0}+P\Delta X_0) +P_RB_1 \,.
\eeqn 
To minimize the first term on the r.h.s. of \eqref{eq-cC-upperbd-1}, one might want to try to solve
\eqn 
	W_2 S =Y^{ext}
\eeqn 
where
\eqn 
	S:= P_R R (\overline{X_0}+P\Delta X_0 ) 
	+P_R B_1 \in \R^{M\times N}
\eeqn 	
has rank $Q < M$ (due to $P_R$). However, this implies that the $QM$ unknown matrix components of $W_2\in\R^{Q\times M}$ are determined by $QN> QM$ equations. This is an overdetermined problem which generically has no solution. However, we can solve the problem to zeroth order in $\Delta X_0$, 
\eqn
	W_2 S_0 =Y^{ext}
\eeqn 
with 
\eqn
	S_0:=S|_{\Delta X_0=0}=P_RR\overline{X_0} = RP \overline{X_0} 
\eeqn 
(recalling from \eqref{eq-ovlnX0-def-1} that $\overline{X_0}$ only contains $Q$  distinct column vectors $\overline{x_{0,j}}$ and their identical copies) and subsequently minimize the resulting expression in the first term on the r.h.s. of \eqref{eq-cC-upperbd-1} with respect to $P_R B_1$.

This is equivalent to requiring that $W_2\in\R^{Q\times M}$ solves
\eqn 
	W_2 R P\overline{X_0^{red}} = Y \,.
\eeqn 
%in order to minimize the first term on the r.h.s. of \eqref{eq-cC-upperbd-1} to zeroth order in $\Delta X_0$.
To this end, we make the ansatz 
\eqn
	W_2=A (R\overline{X_0^{red}})^T
\eeqn 
for some $A\in\R^{Q\times Q}$. Then, recalling that $P\overline{X_0^{red}}=\overline{X_0^{red}}$,
\eqn 
	A (R\overline{X_0^{red}})^T(R\overline{X_0^{red}}) =
	A (\overline{X_0^{red}})^T(\overline{X_0^{red}}) = Y \,,
\eeqn 
so that solving for $A$, we get
\eqn 
	W_2 &=& Y ((\overline{X_0^{red}})^T\overline{X_0^{red}})^{-1}
	(R\overline{X_0^{red}})^T 
	\nonumber\\
	&=&
	\widetilde W_2 R^T
	\,
\eeqn 
with $\widetilde W_2$ as defined in \eqref{eq-tildW-def-1}.
Hereby, $W_2$ is fully determined by $Y$ and $\overline{X_0^{red}}$ (as $\overline{X_0^{red}}$ determines $P$ and $R$).

Next, we can choose $P^\perp_R b_1$ in a suitable manner that
\eqn\label{eq-PperpDeltX0-stop-1}
	\sigma( R P^\perp  \Delta X_0+P^\perp_R B_1) = 0 \,.
\eeqn 
This can be accomplished with 
\eqn 
	P^\perp_R b_1 = - \delta P^\perp_R u_M\in\R^M
\eeqn  
where $\delta$ is defined in \eqref{eq-delt-1}, and
%, and for an arbitrary constant $\mu\geq1$. 
\eqn\label{def-E-u-1}
	u_M:=(1,1,\dots,1)^T \in\R^M \,.
\eeqn
With
\eqn 
	E:=u_Mu_N^T=[u_M\cdots u_M] \in\R^{M\times N} \,,
\eeqn  
this ensures that the argument on the l.h.s. of \eqref{eq-PperpDeltX0-stop-1}, given by 
\eqn 
	 R P^\perp \Delta X_0+P^\perp_R B_1
	=P^\perp_R R \Delta X_0-\delta P^\perp_R E \,,
\eeqn 
is a matrix all of whose components are $\leq0$. Therefore, $\sigma$ maps it to $0$.

Consequently, we find
\eqn 
	\sqrt{N}\cC[W_j,b_j] \leq  
	\|W_2(RP\Delta X_0 +P_RB_1)+B_2\|_{\cL^2}  
	\,.
\eeqn 
To minimize the r.h.s., we may rewrite the bias
\eqn\label{eq-b2-bias-1}
	b_2 \rightarrow b_2 - W_2 P_Rb_1
\eeqn
so that it remains to minimize the r.h.s. of
\eqn 
	\sqrt{N}\cC[W_j,b_j] \leq  
	\|W_2RP\Delta X_0 +B_2\|_{\cL^2}  
	\,.
\eeqn 
with respect to  
\eqn
	B_2 = [b_2\cdots b_2] = b_2 u_N^T \;\; \in\R^{Q\times N} \,.
\eeqn  
We find 
\eqn\label{eq-cC-minimize-1}
	0&=&
	\partial_{b_2}\|W_2 RP\Delta X_0 +\beta_2 u_N^T\|_{\cL^2}^2
	\nonumber\\
	&=&
	\partial_{b_2}\Big(2\tr(W_2RP
	\underbrace{\Delta X_0 u_N }_{=0} b_2^T )
	+\tr( u_N
	\underbrace{b_2^T b_2}_{=|b_2|^2} u_N^T )\Big)\,,
\eeqn 
using cyclicity of the trace.
The first term inside the bracket on the second line is zero because for every $i\in\{1,\dots,M\}$,
\eqn 
	( \Delta X_0 u_N )_i &=& \sum_{\ell=1}^N  [\Delta X_0]_{i\ell}
	\nonumber\\
	&=&\sum_{j=1}^Q \sum_{\ell=1}^{N_j} ( x_{0,j,\ell}-\overline{x_{0,j}})_i
	\nonumber\\
	&=&0\,.
\eeqn 
The second term inside the bracket in \eqref{eq-cC-minimize-1} is proportional to $|b_2|^2$, so that its gradient is proportional to $b_2$, and vanishes for $b_2=0$. Hence, $B_2=0$, which means that
\eqn\label{eq-PRB1-1}
	B_2 = - W_2 P_R B_1
\eeqn 
for the original variable $b_2$ prior to the transformation \eqref{eq-b2-bias-1}.
%where the row-wise average $Av(A)$ of a matrix $A=[a_{ij}]\in\R^{M\times N}$ is given by 
%\eqn\label{eq-Av-def-1}
%	[Av(A) ]_{ij} := 
%	\frac1N\sum_{j=1}^N a_{ij} 
%\eeqn 
%for every $i\in\{1,\dots,Q\}$.
%This follows from $Av(\Delta X_0)=0$, which is an immediate consequence of the definition of $\Delta X_0$.
%In particular, this is consistent with \eqref{eq-PRB1-req-1}.

We thus arrive at an upper bound 
\eqn\label{eq-cC-bd-1-1}
	\min_{W_j,b_j}\sqrt{N}\cC[W_j,b_j] &\leq&  
	\|\widetilde W_2 P\Delta X_0  \|_{\cL^2}
	\nonumber\\
	&=&
	\|Y \Pen[\overline{X_0^{red}}] P\Delta X_0  \|_{\cL^2} \,.
\eeqn  
This establishes the asserted upper bound in \eqref{eq-cC-uppbd-thm-1}.

Next, we bound 
\eqn
	\|\widetilde W_2 P\Delta X_0  \|_{\cL^2}
	&\leq&\|Y \|_{op}\| \Pen[\overline{X_0^{red}}]P\Delta X_0  \|_{\cL^2} \,.
\eeqn 
%Writing
%\eqn
%	P_R=\sum_{j=1}^M \nu_j e_j e_j^T
%	\;\;,\;\;
%	\nu_j\in\{0,1\}
%\eeqn 
%where $\{e_j\}_{j=1}^M$ are the Euclidean orthonormal basis column vectors in $\R^M$ for our given coordinate system, and $Q$ of the coefficients $\nu_j$ equal $1$, while the others are zero, so that $\sum_j\nu_j=Q$.
Let for notational brevity $A:=\Pen[\overline{X_0^{red}}]P\Delta X_0 $. We have
\eqn 
	\| \Pen[\overline{X_0^{red}}]P\Delta X_0  \|_{\cL^2}^2 
	&=&
	\tr(AA^T)
	\nonumber\\
	&=&
	\tr(A^TA)
	\nonumber\\
%	&=&
%	\sum_j \nu_j |(RP\Delta X_0)^T e_j|^2_{\R^N}
%	\nonumber\\
	&=&
	\sum_{\ell=1}^Q\sum_{i=1}^{N_j}
	|\Pen[\overline{X_0^{red}}]P\Delta x_{0,\ell,i} |^2
	\nonumber\\
	&\leq&\Big(\sup_{i,\ell}
	\left|\Pen[\overline{X_0^{red}}]P\Delta x_{0,\ell,i}\right|^2\Big)
	\sum_{\ell=1}^Q\sum_{i=1}^{N_\ell}1
	\nonumber\\
	&=&N \sup_{i,\ell}
	\left|\Pen[\overline{X_0^{red}}]P\Delta x_{0,\ell,i}\right|^2 \,.
\eeqn  
Therefore,
\eqn
	\min_{W_j,b_j}\sqrt{N}\cC[W_j,b_j] 
	&\leq&\|\widetilde W_2 P\Delta X_0  \|_{\cL^2}
	\nonumber\\
	&\leq&\sqrt{N} \;
	\|Y \|_{op}
	\sup_{i,\ell}
	\left|\Pen[\overline{X_0^{red}}]P\Delta x_{0,\ell,i}\right|
	\nonumber\\
	&\leq&\sqrt{N} \;
	\|Y \|_{op} \delta_P
	\,,
\eeqn 
which proves \eqref{eq-cC-average-bd-1}.
This concludes the proof of Theorem \ref{thm-cC-uppbd-1}.
\qed

\section{Proof of Theorem \ref{thm-cC-uppbd-2}}
\label{sec-prf-thm-cC-uppbd-2}

To prove the exact degenerate local minimum in the special case $M=Q$, we note that in this situation,
\eqn 
	P=\1_{Q\times Q} 
	\;\;\;,\;
	P^\perp=0\,.
\eeqn 
In place of \eqref{eq-cC-upperbd-1}, the cost function is given by
\eqn\label{eq-cC-upperbd-1-1}
	\CostN[W_j,b_j] &=&  
	\|W_2\sigma(W_1(\overline{X_0}+  \Delta X_0) 
	+B_1)+B_2-Y^{ext}\|_{\cL^2_\cN} \,,
\eeqn
defined in \eqref{eq-CostN-def-1-0-0},
which is weighted by the inverse of the block diagonal matrix  
\eqn
	\cN=\diag(N_j \1_{N_j\times N_j} \, | \, j=1,\dots,Q)
	\;\in\R^{N\times N} \,.
\eeqn 
Using a similar choice for $B_1$ as in \eqref{eq-Prb1-choice-1}, 
%and possibly applying a rotation $R\in O(Q)$, 
we may assume that the matrix components of $W_1(\overline{X_0}+\Delta X_0)+B_1$ are non-negative. To be precise, we let
\eqn\label{eq-Thm3.2-b1-1}
	b_1=\beta_1 u_Q %W_1 \overline{x}
\eeqn 
for some $\beta_1\geq 0$ large enough that %the set of translated input vectors 
\eqn\label{eq-beta1-cone-2}
	%\{
	x_{0,j,i}+\beta_1u_Q \in\R_+^Q
	%\overline{x}\in\R^Q\,|\,j=1,\dots,Q \;,\;i=1,\dots,N_j\} \subset \cK_{\omega,\theta}
\eeqn 
for all $j=1,\dots,Q$, $i=1,\dots,N_j$.
%%is contained in the interior of a cone 
%\eqn
%	\cK_{\omega,\theta} = \{x\in\R^Q \, | \, |(x,\omega)|>\cos\frac\theta2\}
%\eeqn
%of opening angle $\theta<\frac{\pi}{2}-\eta$ where $\eta>\delta$, and centered in the direction of a unit vector $\omega\in\R^Q$, $|\omega|=1$.
%Then, we may choose $R\in O(Q)$ suitably to rotate $\cK_{\omega,\theta}$ into $\R_+^Q\subset\R^Q$, which is for instance accomplished if $R\omega$ is parallel to the diagonal, $(1,1,\dots,1)^T\in\R^Q$.
This is achieved with 
\eqn
	\beta_1 \geq 2\max_{j,i}|x_{0,j,i}| \,.
\eeqn
Hence, choosing 
\eqn\label{eq-Thm3.2-W1-1}
	W_1=\1_{Q\times Q}\,,
\eeqn 
we find 
\eqn 
	\sigma(W_1 X_0+B_1) 
	%&=&  \sigma(R( X_0+\beta_1[\overline{x}\cdots\overline{x}]))
	%\nonumber\\
	%&=& R( X_0+\beta_1[\overline{x}\cdots\overline{x}])
	%\nonumber\\
	%&=& 
	= \overline{X_0}+  \Delta X_0
	+B_1 \,.
\eeqn
Thus, \ref{eq-cC-upperbd-1-1} implies
\eqn\label{eq-cC-upperbd-1-2}
	\min_{W_j,b_j}\CostN[W_j,b_j] \leq 
	\|W_2(\overline{X_0}+  \Delta X_0) 
	+W_2B_1+B_2-Y^{ext}\|_{\cL^2_\cN} \,.
\eeqn
To minimize the r.h.s., we note that since $W_2$ and $B_2$ are both unknown, we can redefine $W_2\rightarrow W_2$, and we can bias 
\eqn\label{eq-b2-bias-1-2}
 	b_2\rightarrow b_2 - W_2 b_1 \,.
\eeqn 
Therefore, we find
\eqn\label{eq-cC-upperbd-1-2.1}
	\min_{W_j,b_j}\CostN[W_j,b_j] \leq   \min_{W_2,b_2}
	\|W_2(\overline{X_0}+  \Delta X_0) 
	+B_2-Y^{ext}\|_{\cL^2_\cN} \,.
\eeqn
Minimizing with respect to $W_2\in\R^{Q\times Q}$ is a least squares problem whose solution has the form
\eqn\label{eq-leastsq-1}
	(W_2 (\overline{X_0}+  \Delta X_0)-Y^{ext}_{b_2})\cP = 0 \,,
\eeqn 
where 
\eqn 
	Y^{ext}_{b_2} := Y^{ext}-B_2 \,,
\eeqn 
$\cP^T$ is a projector onto the range of $X_0^T$, and remains to be determined.
To solve \eqref{eq-leastsq-1}, we require $W_2$ to satisfy
\eqn\label{eq-W2-QMeq-1}
	\lefteqn{
	W_2 (\overline{X_0}+  \Delta X_0)\cN^{-1}(\overline{X_0}+  \Delta X_0)^T 
	}
	\nonumber\\
	&&\hspace{2cm}= Y^{ext}_{b_2}\cN^{-1}(\overline{X_0}+  \Delta X_0)^T 
	\;\; \in\R^{Q\times Q} \,.
\eeqn 
Because
\eqn 
	\overline{X_0}&=&[\overline{x_{0,1}}u_{N_1}^T\cdots \overline{x_{0,Q}}u_{N_Q}^T]
	\,,
	\nonumber\\
	Y^{ext}_{b_2}
	&=&[(y_1-b_2) u_{N_1}^T\cdots (y_Q-b_2)u_{N_Q}^T] \,,
\eeqn 
where 
\eqn
	u_{N_j}:=(1,1,\dots,1)^T \;\; \in\R^{N_j} \,,
\eeqn 
we find that
\eqn 
	\Delta X_0\cN^{-1}(\overline{X_0})^T
	&=&
	\sum_{j=1}^Q \frac1{N_j}\Delta X_{0,j} u_{N_j} \overline{x_{0,j}}^T 
	\nonumber\\
	&=& \sum_{j=1}^Q 
	\underbrace{\Big(\frac1{N_j}\sum_{i=1}^{N_j} \Delta x_{0,j,i}\Big)}_{=0} \overline{x_{0,j}}^T
	\nonumber\\
	&=&0 \,.
\eeqn 
For the same reason,
\eqn 
	\overline{X_0}\cN^{-1}\Delta X_0^T = 0
	\;\;\;,\;
	Y^{ext}\cN^{-1}\Delta X_0^T = 0 \,.
\eeqn 
Moreover, 
\eqn
	Y^{ext}_{b_2}\cN^{-1}\overline{X_0}^T &=& 
	\sum_{j=1}^Q \frac1{N_j}(y_j-b_2) \underbrace{u_{N_j}^T u_{N_j} }_{= N_j}\overline{x_{0,j}}^T 
	\nonumber\\
	&=&
	Y_{b_2} \overline{X_0^{red}}^T \,,
\eeqn
where
\eqn
	Y_{b_2}
	=[(y_1-b_2) \cdots (y_Q-b_2)] \;\;\in\R^{Q\times Q}
\eeqn 
and 
\eqn 
	\overline{X_0}\cN^{-1}\overline{X_0}^T = \overline{X_0^{red}}\,\overline{X_0^{red}}^T \,.
\eeqn 
Therefore, \eqref{eq-W2-QMeq-1} reduces to
\eqn\label{eq-W2-QMeq-2}
	W_2 (\overline{X_0^{red}}\,\overline{X_0^{red}}^T+   
	\Delta X_0\cN^{-1}\Delta X_0^T) 
	= Y_{b_2}\overline{X_0^{red}}^T \,, 
\eeqn 
and invertibility of $\overline{X_0^{red}}\,\overline{X_0^{red}}^T\in\R^{Q\times Q}$ implies invertibility of the matrix in brackets on the l.h.s. of \eqref{eq-W2-QMeq-2}, because
\eqn\label{eq-D2-def-1}
	D_2[X_0]:=\Delta X_0\cN^{-1}\Delta X_0^T = 
	\sum_{j=1}^Q \frac1{N_j} \sum_{i=1}^{N_j}
	\Delta x_{0,j,i} \Delta x_{0,j,i}^T \;\;\in\R^{Q\times Q}\,
\eeqn 
is a non-negative operator. 
Hence, 
\eqn\label{eq-W2-QMeq-3}
	W_2 
	= Y_{b_2}\overline{X_0^{red}}^T(\overline{X_0^{red}}\,\overline{X_0^{red}}^T+   
	\Delta X_0\cN^{-1}\Delta X_0^T)^{-1} \,,
\eeqn 
so that
\eqn 
	W_2X_0=Y^{ext}_{b_2}\cP \,.
\eeqn 
Here,
\eqn\label{eq-cP-def-1}
	\cP:=\cN^{-1}X_0^T(X_0\cN^{-1}X_0^T)^{-1}X_0 \in\R^{N\times N}
\eeqn 
is a rank $Q$ projector $\cP^2=\cP$ which is orthogonal with respect to the inner product on $\R^{N}$ defined by $\cN$, i.e., $\cP^T\cN=\cN\cP$. $\cP^T$ is the projector onto the range of $X_0^T$. We note that with respect to the inner product defined with $\cN^{-1}$ (which we are using here), we have $\cN^{-1}\cP^T=\cP\cN^{-1}$.

Next, in order to control $W_2X_0-Y^{ext}_{b_2}=-Y^{ext}_{b_2}\cP^\perp$, we observe that
\eqn 
	Y^{ext}_{b_2}\cP&=&Y^{ext}_{b_2}\cN^{-1}X_0^T(X_0\cN^{-1}X_0^T)^{-1}X_0
	\nonumber\\
	&=&Y_{b_2}\overline{X_0^{red}}^T
	(\overline{X_0^{red}}\,\overline{X_0^{red}}^T+D_2[X_0])^{-1}
	X_0
	\nonumber\\
	&=&Y_{b_2}\overline{X_0^{red}}^T
	(\overline{X_0^{red}}\,\overline{X_0^{red}}^T)^{-1}
	X_0
	\nonumber\\
	&&- Y_{b_2}\overline{X_0^{red}}^T 
	(\overline{X_0^{red}}\,\overline{X_0^{red}}^T)^{-1} 
	D_2[X_0]
	(X_0\cN^{-1}X_0^T)^{-1}
	X_0
	\nonumber\\
	&=&Y_{b_2}
	(\overline{X_0^{red}})^{-1}
	X_0
	\\
	&&- Y_{b_2}	
	(\overline{X_0^{red}})^{-1} 
	D_2[X_0]
	(X_0\cN^{-1}X_0^T)^{-1}
	X_0
	\nonumber 
\eeqn 
where we have used the matrix identity
\eqn
	(A+B)^{-1}&=&A^{-1}-A^{-1}B(A+B)^{-1}
	\nonumber\\
	&=&A^{-1}-(A+B)^{-1}BA^{-1}
\eeqn 
for $A$, $A+B$ invertible.
We observe that
\eqn 
	Y_{b_2}(\overline{X_0^{red}})^{-1}
	X_0 
	&=&
	Y_{b_2}(\overline{X_0^{red}})^{-1}\overline{X_0}  
	+ Y_{b_2}(\overline{X_0^{red}})^{-1}
	\Delta{X_0}
	\nonumber\\
	&=&
	Y^{ext}_{b_2} + Y_{b_2}(\overline{X_0^{red}})^{-1}
	\Delta{X_0} \,.
\eeqn
To pass to the last line, we used that $(\overline{X_0^{red}})^{-1}\overline{X_0}=\,[e_1 u_{N_1}^T \cdots e_Q u_{N_Q}^T]$ where $\{e_j \in\R^Q \,| \, j=1,\dots, Q\}$ are the unit basis vectors.

We therefore conclude that, due to $\cP^\perp=\1-\cP$,
\eqn
	Y^{ext}_{b_2}\cP^\perp&=&-Y_{b_2}(\overline{X_0^{red}})^{-1}
	\Delta{X_0} + \cR 
\eeqn
where
\eqn\label{eq-cR-def-1}
	\cR 
	&:=& Y_{b_2}	
	(\overline{X_0^{red}})^{-1} 
	D_2[X_0]
	(X_0\cN^{-1}X_0^T)^{-1}
	X_0  
\eeqn 
satisfies
\eqn\label{eq-cR-cRP-1-0}
	\cR\cP = \cR \,,
\eeqn 
as can be easily verified.

At this point, we have found that
\eqn\label{eq-cC-upperbd-1-3}
	\min_{W_j,b_j}\cC_{\cN}[W_j,b_j] \leq \min_{b_2}   
	\|Y^{ext}_{b_2}\cP^\perp\|_{\cL^2_{\cN}} \,.
\eeqn
To minimize with respect to $b_2$, we set
\eqn 
	0&=&\partial_{b_2} \|Y^{ext}_{b_2}\cP^\perp\|_{\cL^2_{\cN}} ^2
	\nonumber\\
	&=&
	\partial_{b_2}\Big(
	-2\tr\Big(Y^{ext}\cP^{\perp}\cN^{-1}(\cP^{\perp})^T u_N b_2^T \Big)
	+ \tr\big(b_2 u_N^T \cP^\perp\cN^{-1}(\cP^\perp)^T u_N b_2^T\big) \Big)
	\nonumber\\
	&=&
	\partial_{b_2}\Big(
	-2 \tr\Big(
	b_2^T Y^{ext}\cP^{\perp}\cN^{-1}(\cP^{\perp})^T u_N \Big)
	+ |\cN^{-1/2}(\cP^\perp)^T u_N |^2 |b_2|^2 \Big)
	\nonumber\\
	&=&
	\partial_{b_2}\Big(
	-2 \tr\Big( b_2^T Y_{b_2}(\overline{X_0^{red}})^{-1}
	\underbrace{\Delta{X_0} \cN^{-1} u_N }_{=0}\Big)
	\;+\; 
	\tr\Big(
	b_2^T \underbrace{\cR \cN^{-1}(\cP^{\perp})^T}_{=\cR\cP\cN^{-1}\cP^\perp=0} u_N 
	\Big)\nonumber\\
	&&\hspace{2cm}
	+ |\cN^{-1/2}(\cP^\perp)^T u_N |^2 |b_2|^2 \Big)
\eeqn 
using $\cR=\cR\cP$ from \eqref{eq-cR-cRP-1-0} so that $\cR\cN^{-1}\cP^\perp=\cR\cP\cN^{-1}\cP^\perp=\cR\cP\cP^\perp\cN^{-1}=0$, and recalling $u_N=(1,1,\dots,1)^T\in\R^N$. This implies that 
\eqn
	b_2=0 \,.
\eeqn
We note that this corresponds to 
\eqn\label{eq-Thm3.2-b2-1}
	b_2 = - W_2 b_1
\eeqn 
prior to the bias \eqref{eq-b2-bias-1-2}, and hence arrive at \eqref{eq-Thm3.2-b1b2-1}, as
claimed.

Next, we derive a simplified expression for $\|Y^{ext}\cP^\perp\|_{\cL^2_{\cN}}^2$.
For notational convenience, let
\eqn
	\Delta_1^{rel} &:=& (\overline{X_0^{red}})^{-1} 
	\Delta X_0 
	\nonumber\\
	\Delta_2^{rel} &:=& (\overline{X_0^{red}})^{-1} 
	D_2[X_0](\overline{X_0^{red}})^{-T}
	\nonumber\\
	&=& \Delta_1^{rel}  \cN^{-1}(\Delta_1^{rel} )^{T}
\eeqn 
where we recall \eqref{eq-D2-def-1}.
Then, we obtain (with $b_2=0$)
\eqn 
	\cR &=& Y \Delta_2^{rel}\big(1+\Delta_2^{rel}\big)^{-1}
	(\overline{X_0^{red}})^{-1} X_0
	\nonumber\\
	&=&Y \Delta_2^{rel}\big(1+\Delta_2^{rel}\big)^{-1}
	\Big([e_1u_{N_1}^T\cdots e_Q u_{N_Q}^T]+ \Delta_1^{rel} \Big)
\eeqn 
where we used  $X_0\cN^{-1}X_0^T=\overline{X_0^{red}}(1+\Delta_2^{rel})(\overline{X_0^{red}})^T$ in \eqref{eq-cR-def-1} to obtain the first line.
We find that
\eqn 
	\lefteqn{
	-Y^{ext}\cP^\perp
	}
	\nonumber\\ 
	&=&
	Y\Delta_1^{rel} - Y \Delta_2^{rel}\big(1+\Delta_2^{rel}\big)^{-1}
	\Big([e_1u_{N_1}^T\cdots e_Q u_{N_Q}^T]+ \Delta_1^{rel} \Big)
	\nonumber\\
	&=&
	Y(1-\Delta_2^{rel}\big(1+\Delta_2^{rel}\big)^{-1})\Delta_1^{rel} 
	- Y\Delta_2^{rel}\big(1+\Delta_2^{rel}\big)^{-1} 
	[e_1u_{N_1}^T\cdots e_Q u_{N_Q}^T]
	\nonumber\\
	&=&
	Y(1+\Delta_2^{rel})^{-1}\Delta_1^{rel} 
	- Y\Delta_2^{rel}\big(1+\Delta_2^{rel}\big)^{-1} 
	[e_1u_{N_1}^T\cdots e_Q u_{N_Q}^T] \,.
\eeqn 
Therefore, we obtain that 
\eqn  
	\|Y^{ext}\cP^\perp\|_{\cL^2_{\cN}}^2  
	&=&(I)+(II)+(III)
\eeqn 
where
\eqn 
	(I)&:=&\tr\Big( Y(1+\Delta_2^{rel})^{-1}\Delta_1^{rel} \cN^{-1}
	(\Delta_1^{rel} )^T(1+\Delta_2^{rel})^{-T}Y^T\Big)
	\nonumber\\
	&=&
	\tr\Big( Y(1+\Delta_2^{rel})^{-1}\Delta_2^{rel} (1+\Delta_2^{rel})^{-1}Y^T\Big)
\eeqn 
where we note that $\Delta_2^{rel}=(\Delta_2^{rel})^T\in\R^{Q\times Q}$ is symmetric.
Moreover,
\eqn 
	(II)&:=&-2\tr\Big( Y(1+\Delta_2^{rel})^{-1}
	\Delta_1^{rel} \cN^{-1}
	[e_1u_{N_1}^T\cdots e_Q u_{N_Q}^T]^T 
	\big(1+\Delta_2^{rel}\big)^{-1} \Delta_2^{rel}
	Y^T\Big)
	\nonumber\\
	&=&-\sum_{j=1}^Q 2\tr\Big( Y(1+\Delta_2^{rel})^{-1} (\overline{X_0^{red}})^{-1}
	\underbrace{\frac{1}{N_j}\Delta X_{0,j} u_{N_j} e_j^T }_{=0}
	\big(1+\Delta_2^{rel}\big)^{-1} \Delta_2^{rel}
	Y^T\Big)
	\nonumber\\
	&=&0
\eeqn 
where we used that $\frac{1}{N_j}\Delta X_{0,j} u_{N_j} =\overline{\Delta X_{0,j}}=0$ for all $j=1,\dots,Q$.
Furthermore,
\eqn 
 	(III)&:=& 
 	\tr\Big(
 	Y\Delta_2^{rel}\big(1+\Delta_2^{rel}\big)^{-1} 
	[e_1u_{N_1}^T\cdots e_Q u_{N_Q}^T]\cN^{-1}[e_1u_{N_1}^T\cdots e_Q u_{N_Q}^T]^T
	\nonumber\\
	&&\hspace{2cm}
	\big(1+\Delta_2^{rel}\big)^{-1} \Delta_2^{rel}
	Y^T
 	\Big)
 	\nonumber\\
 	&=&
 	\tr\Big(
 	Y\Delta_2^{rel}\big(1+\Delta_2^{rel}\big)^{-1} 
 	\sum_{j=1}^Q\underbrace{\frac{1}{N_j} u_{N_j}^T u_{N_j} }_{=1}
 	e_je_j^T 
	\big(1+\Delta_2^{rel}\big)^{-1} \Delta_2^{rel}
	Y^T
 	\Big)
 	\nonumber\\
 	&=&
 	\tr\Big(
 	Y\Delta_2^{rel}\big(1+\Delta_2^{rel}\big)^{-1} 
	\big(1+\Delta_2^{rel}\big)^{-1} \Delta_2^{rel}
	Y^T
 	\Big)
 	\nonumber\\
 	&=&
 	\tr\Big(Y
 	\big(1+\Delta_2^{rel}\big)^{-1} (\Delta_2^{rel})^2
	\big(1+\Delta_2^{rel}\big)^{-1}  
	Y^T
 	\Big)
\eeqn
using $\sum_{j=1}^Q e_je_j^T =\1_{Q\times Q}$, and commutativity of $\Delta_2^{rel}$ and $(1+\Delta_2^{rel})^{-1}$.

We conclude that
\eqn\label{eq-CostN-YPperp-1-0}
	\|Y^{ext}\cP^\perp\|_{\cL^2_{\cN}}^2  
	&=&(I)+(III)
	\nonumber\\
	&=&
	\tr\Big(
 	Y\big(1+\Delta_2^{rel}\big)^{-1} 
 	\Delta_2^{rel}(1+\Delta_2^{rel})
	\big(1+\Delta_2^{rel}\big)^{-1}  
	Y^T
 	\Big)
 	\nonumber\\
	&=&
	\tr\Big(
 	Y\Delta_2^{rel}\big(1+\Delta_2^{rel}\big)^{-1}  
	Y^T
 	\Big)
 	\nonumber\\
	&=&
	\tr\Big(
 	Y|\Delta_2^{rel}|^{\frac12}\big(1+\Delta_2^{rel}\big)^{-1}  
	|\Delta_2^{rel}|^{\frac12}Y^T
 	\Big)
 	\nonumber\\
 	&\leq&\|\big(1+\Delta_2^{rel}\big)^{-1}\|_{op} \; \tr\Big(
 	Y\Delta_2^{rel} Y^T
 	\Big)
\eeqn
since the matrix $\Delta_2^{rel}\geq0$ is positive semidefinite.
We note that
\eqn
	\tr\Big(
 	Y\Delta_2^{rel} Y^T
 	\Big) 
 	&=&\tr\Big(
 	Y\Delta_1^{rel}\cN^{-1}(\Delta_1^{rel})^T Y^T
 	\Big)
	\nonumber\\
	&=&
	\|Y(\overline{X_0^{red}})^{-1}\Delta{X_0}\|_{\cL^2_{\cN}}^2 \,.
\eeqn
Moreover, letting $\lambda_-:=\inf {\rm spec} (\Delta_2^{rel})$ and $\lambda_+:=\sup {\rm spec} (\Delta_2^{rel})$ denote the smallest and largest eigenvalue of the non-negative, symmetric matrix $\Delta_2^{rel}\geq0$,  
\eqn 
	0\leq \lambda_- \leq \lambda_+ =\|\Delta_2^{rel}\|_{op} =
	\|\Delta_1^{rel}\cN^{-1}(\Delta_1^{rel})^T\|_{op} \leq C\delta_P^2 \,.
\eeqn 
Therefore, 
\eqn 
	\|\big(1+\Delta_2^{rel}\big)^{-1}\|_{op} 
	= ( 1+  \lambda_- )^{-1}
	\leq 1 - C_0'\delta_P^2
\eeqn
for a constant $C_0'=0$ if $\rank(\Delta_2^{rel} )<Q$ whereby $\lambda_-=0$, and $C_0'>0$ if $\rank(\Delta_2^{rel} )=Q$.

We find the following bound from \eqref{eq-CostN-YPperp-1-0},
\eqn\label{eq-cC-upperbd-1-4} 
	\|Y^{ext}\cP^\perp\|_{\cL^2_{\cN}} 
 	&\leq&\sqrt{
 	\|\big(1+\Delta_2^{rel}\big)^{-1}\|_{op} \; \tr\Big(
 	Y\Delta_2^{rel} Y^T
 	\Big) }
 	\nonumber\\
	&\leq& (1-C_0\delta_P^2)
	\; \|Y(\overline{X_0^{red}})^{-1}\Delta{X_0}\|_{\cL^2_{\cN}} 
	\,,
\eeqn
as claimed, where $C_0$ is proportional to $C_0'$.

Finally, we use that
\eqn 
	\widetilde W_2 &=& Y\overline{X_0^{red}}^T (\overline{X_0^{red}}\,\overline{X_0^{red}}^T)^{-1}
	\nonumber\\
	&=&Y  (\overline{X_0^{red}})^{-1} \,
\eeqn 
agrees with \eqref{eq-tildW-def-1} because $\overline{X_0^{red}}$ is invertible, and
hence $(\overline{X_0^{red}})^{-1}=\Pen[\overline{X_0^{red}}]$. Hence, we observe that
\eqref{eq-W2-QMeq-3} implies  
\eqn 
	W_2 
	&=& \widetilde W_2  -\widetilde W_2  
	(\Delta X_0\cN^{-1}\Delta X_0^T) 
	(\overline{X_0^{red}}\,\overline{X_0^{red}}^T+   
	\Delta X_0\cN^{-1}\Delta X_0^T)^{-1}
	\nonumber\\
	&=& \widetilde W_2  - Y (\overline{X_0^{red}})^{-1}
	(\Delta X_0\cN^{-1}\Delta X_0^T) (\overline{X_0^{red}})^{-T}
	\nonumber\\
	&&\hspace{2cm}
	\Big(\1+ (\overline{X_0^{red}})^{-1}  
	(\Delta X_0\cN^{-1}\Delta X_0^T)
	(\overline{X_0^{red}})^{-T}\Big)^{-1}
	(\overline{X_0^{red}})^{-1}
	\nonumber\\
	&=& \widetilde W_2  - Y \Delta_2^{rel}
	(\1+ \Delta_2^{rel})^{-1}
	(\overline{X_0^{red}})^{-1}
\eeqn
Therefore, in operator norm,
\eqn 
	\|W_2-\widetilde W_2 \|_{op}
	&\leq&\|Y\|_{op}\|\Delta_2^{rel}\|_{op}\|(\1+ \Delta_2^{rel})^{-1}\|_{op}\|(\overline{X_0^{red}})^{-1}\|_{op}
	\nonumber\\
	&\leq & C \delta_P^2
\eeqn 	
where we used $\|Y\|_{op}<C$, $\|\Delta_2^{rel}\|_{op}<C\delta_P^2$, $\|(\1+ \Delta_2^{rel})^{-1}\|_{op}\leq1$, $\|(\overline{X_0^{red}})^{-1}\|_{op}<C$ to obtain the last line.

We note that when $M>Q$, it follows that $\overline{X_0^{red}}(\overline{X_0^{red}})^T$ is an $M\times M$ matrix of rank $Q$, and is hence not invertible, so that the above arguments do not apply.

From here on, we denote the weights and biases determined above in \eqref{eq-Thm3.2-W1-1}, \eqref{eq-W2-QMeq-3} and in \eqref{eq-Thm3.2-b1-1}, \eqref{eq-Thm3.2-b2-1}, by $W_1^*,W_2^*$ and $b_1^*,b_2^*$, respectively. 
To prove that $\cC_{\cN}[W_i^*,b_i^*]$ is a local minimum of the cost function $\CostN$, we observe that for 
\eqn
	\rho:=\max_{j,i} |x_{0,j,i}|
\eeqn
and 
\eqn 
	\beta_1\geq 2\rho 
	\;\;\;,\;\;
	b_1^*=\beta_1 u_Q\,,
\eeqn 
we have
\eqn 
	W_1^* x_{0,j,i} + b_1^*&=&x_{0,j,i} + b_1^* 
	\nonumber\\
	&\in& B_\rho(\beta_1 u_Q) \subset \R_+^Q \,.
\eeqn 
$\forall j=1,\dots,Q$ and $\forall\;i=1,\dots,N_j $.
That is, all training input vectors are contained in a ball of radius $\rho$ centered at a point on the diagonal $\beta_1 u_Q\in\R^Q$ with $\beta_1\geq2\rho$. This ensures that the coordinates of all $W_1^* x_{0,j,i} +b_1^*$ are strictly positive, and therefore,
\eqn\label{eq-posquadr-1}
	\sigma(W_1^* X_0+B_1^*) = W_1^* X_0+B_1^* \,.
\eeqn  
For $\epsilon >0$ small, we consider an infinitesimal translation
\eqn
 	b_1^* \rightarrow b_1^*+\widetilde b_1
\eeqn 
combined with an infinitesimal transformation 
\eqn 
	W_1^*\rightarrow W_1^*  \widetilde W_1
\eeqn 
where $|\widetilde b_1|<\epsilon$ and $\|\widetilde W_1-\1_{Q\times Q}\|_{op}<\epsilon$. Then, we have that
\eqn\label{eq-W1-infinitbias-1}
	\lefteqn{
	W_1^*  \widetilde W_1 x_{0,j,i} + b_1^*+\widetilde b_1
	}
	\nonumber\\
	&=&\underbrace{W_1^*   x_{0,j,i} + b_1^*}_{\in B_\rho(\beta_1 u_Q)} + 
	W_1^*  (\widetilde W_1 -\1_{Q\times Q}) x_{0,j,i} + \widetilde b_1
\eeqn
where
\eqn 
	|W_1^*  (\widetilde W_1 -\1_{Q\times Q}) x_{0,j,i} + \widetilde b_1|
	&\leq& \|W_1^*\|_{op} \|\widetilde W_1 -\1_{Q\times Q}\|_{op} |x_{0,j,i}| + |\widetilde b_1|
	\nonumber\\
	&\leq& \|W_1^*\|_{op} \; \epsilon \; \max_{j,i}|x_{0,j,i}| + \epsilon 
	\nonumber\\
	&\leq& C \epsilon 
\eeqn 
for some constant $C$ independent of $j,i$.
Therefore, we conclude that \eqref{eq-W1-infinitbias-1} is contained in $B_{\rho+C\epsilon}(\beta_1u_Q)$, which lies in the positive sector $\R_+^Q\subset\R^Q$ for sufficiently small $\epsilon>0$. But this implies that 
\eqn 
	\sigma(W_1^*\widetilde W_1 X_0+B_1^*+\widetilde B_1) = 
	W_1^*\widetilde W_1 X_0+B_1^*+\widetilde B_1 \,.
\eeqn  
Thus, all arguments leading to the result in \eqref{eq-Thm3.2-CostN-1-0} stating that  
\eqn\label{eq-CostN-expl-1-1}
	\CostN[W_i^*,b_i^*]
	=
	\|Y^{ext}\cP^{\perp}\|_{\cL^2_{\cN}}  \,,
\eeqn 
for the choice of weights and biases $W_i^*,b_i^*$, equally apply to the infinitesimally deformed $W_1^*\widetilde W_1$, $b_1^*+\widetilde b_1$ and the corresponding expressions for $W_2^*\widetilde W_2$, $b_2^*+\widetilde b_2$.

We now make the crucial observation that the right hand side of \eqref{eq-CostN-expl-1-1} does not depend on $W_i^*,b_i^*$, as $\cP^\perp$ only depends on the training inputs $X_0$; see \eqref{eq-cP-def-1}.

Therefore,
\eqn 
	\CostN[W_i^*,b_i^*] = \CostN[W_i^*\widetilde W_i,b_i^*+\widetilde b_i] \,,
\eeqn 
and the variation in the perturbations $\widetilde W_i,\widetilde b_i$ is zero.
This implies that $\CostN[W_i^*,b_i^*] $ is a local minimum of the cost function, as it was obtained from a least square minimization problem. In particular, it is degenerate, and by repeating the above argument, one concludes that it has the same value for all weights and biases $W_i,b_i$ which allow for the condition \eqref{eq-posquadr-1} to be satisfied.  

Finally, we note that the reparametrization of the training inputs $X_0\rightarrow K X_0$, for any arbitrary $K\in GL(Q)$, induces
\eqn 
	\overline{X_0}\rightarrow K\overline{X_0}
	\;\;\;,\;\;
	\Delta X_0\rightarrow K\Delta X_0 \,.
\eeqn 
In particular, this implies that
\eqn
	\Delta_1^{rel} &=& (\overline{X_0^{red}})^{-1} 
	\Delta X_0 
	\nonumber\\
	&\rightarrow&(K\overline{X_0^{red}})^{-1} 
	K\Delta X_0 
	\nonumber\\
	&=&(\overline{X_0^{red}})^{-1} 
	K^{-1}K\Delta X_0 
	\nonumber\\
	&=&
	\Delta_1^{rel}
\eeqn
is invariant under this reparametrization, and hence,
\eqn 
	\Delta_2^{rel}    
	&=& \Delta_1^{rel}  \cN^{-1}(\Delta_1^{rel} )^{T}
\eeqn 
also is.
From the third and fourth lines in \eqref{eq-CostN-YPperp-1-0} follows that $\CostN[W_i^*,b_i^*]$ is a function only of $\Delta_2^{rel} $. We note that the reparametrization invariance of $\cP$ (and thus of $\cP^\perp$) can also be directly verified from \eqref{eq-cP-def-1} by inspection.
This implies that $\CostN[W_i^*,b_i^*]$ is invariant under reparametrizations of the training inputs $X_0\rightarrow K X_0$, for all $K\in GL(Q)$.

This concludes the proof of Theorem \ref{thm-cC-uppbd-2}. 
\qed

\section{Proof of Theorem \ref{thm-DL-geometry-1}}
\label{sec-prf-thm-DL-geometry-1}

Here, we prove Theorem \ref{thm-DL-geometry-1}, and elucidate the geometric meaning of the
constructively trained network obtained in Theorem \ref{thm-cC-uppbd-1}.
% Section \ref{sec-theor-DL-1}.  
Given a test
input $x\in\R^M$, the constructively trained network metrizes the linear subspace
$\ran(P)$ of the input space $\R^M$, and determines to which equivalence class of training
inputs $Px$ is closest. To  elucidate this interpretation, we observe that
\eqn\label{eq-tildW-alg-1}
	\widetilde W_2 = Y ((\overline{X_0^{red}})^T\overline{X_0^{red}})^{-1}
	(\overline{X_0^{red}})^T \in \R^{Q\times M} \,
\eeqn 
implies that
\eqn 
	\widetilde W_2 P = \widetilde W_2
	\;\;\;{\rm and}
	\;
	W_2^* P_R = W_2^* \,
\eeqn 
and hence,
\eqn 
	 \widetilde W_2 \overline{X_0^{red}} = Y \,,
\eeqn 
or in terms of column vectors,
\eqn\label{eq-yj-id-1}
	W_2^* R P\overline{x_{0,j}} =\widetilde W_2 \overline{x_{0,j}} = y_j \,.
\eeqn  
Therefore,
\eqn\label{eq-cCj-uppbd-1}
	\cC_j[x] &=& |W_2^*(W_1^* x + b_1^*)_+ +b_2^*-y_j| 
	\nonumber\\
	&=& |W_2^*(R x + b_1^*)_+ +b_2^*-y_j| 
	\nonumber\\
	&=& |W_2^*(R P x + RP^\perp x + P_Rb_1^* + P^\perp_R b_1^*)_+ +b_2^*-y_j| 
	\nonumber\\
	&=& |W_2^*((P_R R P x+ P_Rb_1^*)_+ 
	+ (P^\perp_R RP^\perp x  + P^\perp_R b_1^*)_+) +b_2^*-y_j| 
	\nonumber\\
	&=& |W_2^*(P_R R P x-y_j) + W_2(P^\perp_R RP^\perp x  -\delta P^\perp_R u)_+| 
	\nonumber\\
	&=& |W_2^*(P_R R P (x-\overline{x_{0,j}})) | 
\eeqn
where by construction of $P$, we have $P\overline{x_{0,j}}=\overline{x_{0,j}}$. Passing to the fourth line, we used $P_R RP=RP$, passing to the fifth line, we recalled $W_2^*P_Rb_1^*+b_2^*=0$, and passing to the sixth line, we used \eqref{eq-yj-id-1}. Moreover, we recall that $u$ is defined in \eqref{def-E-u-1}. 
To pass to the sixth line, we also used that
\eqn 
	P^\perp_R (P^\perp_R RP^\perp x  -\delta P^\perp_R u)_+ 
	= (P^\perp_R RP^\perp x  -\delta P^\perp_R u)_+
\eeqn 
because the vector inside $(\cdot)_+$ only has components in the range of $P^\perp_R$, which is diagonal in the given coordinate system, and because $(\cdot)_+$ acts component-wise. Therefore, since also $W_2^* P_R = W_2^*$, we have
\eqn 
	\lefteqn{
	W_2^*(P^\perp_R RP^\perp x  -\delta P^\perp_R u)_+
	}
	\nonumber\\
	&=& W_2^* P_R P^\perp_R (P^\perp_R RP^\perp x  -\delta P^\perp_R u)_+ 
	\nonumber\\
	&=& 0 \,.
\eeqn 
With $W_2^* P_R R P =\widetilde W_2 P$, \eqref{eq-cCj-uppbd-1} implies that
\eqn
	\cC_j[x] = d_{\widetilde W_2}(Px,\overline{x_{0,j}}) 
\eeqn
where 
\eqn
	d_{\widetilde W_2}(x,y):=|\widetilde W_2 P(x - y)| \,,
\eeqn 
for $x,y\in \ran(P)$, defines a metric in the range of $P$, which is a $Q$-dimensional linear subspace of the input space $\R^M$. 
This is because the matrix $\widetilde W_2 =\widetilde W_2 P$ has full rank $Q$.

We can therefore reformulate the identification of an input $x\in\R^M$ with an output $y_{j^*}$ via the constructively trained shallow network as the solution of the metric minimization problem
\eqn 
	j^* = \argmin_{j\in\{1,\dots,Q\}} (d_{\widetilde W_2}(Px,\overline{x_{0,j}}))
\eeqn 
in the range of $P$. 
\qed

$\;$\\
\noindent
{\bf Acknowledgments:} 
T.C. gratefully acknowledges support by the NSF through the grant DMS-2009800, and the RTG
Grant DMS-1840314 - {\em Analysis of PDE}. P.M.E. was supported by NSF grant DMS-2009800
through T.C.  \\

\bibliographystyle{alpha}
%\bibliography{thomas, /Users/patricia/academic/bib/ml, chatgpt}

\newcommand{\etalchar}[1]{$^{#1}$}

\end{document}